\documentclass[journal]{IEEEtran}
\usepackage{algpseudocode}
\usepackage{amsmath}
\usepackage[linesnumbered,ruled,vlined]{algorithm2e}
\usepackage{comment}
\usepackage{graphicx}
\usepackage{subcaption}

\usepackage{mathtools,amssymb}
\usepackage{amsmath}
\newcommand{\norm}[1]{\left\lVert#1\right\rVert}
\usepackage[linesnumbered,ruled,vlined]{algorithm2e}
\usepackage{algpseudocode}
\usepackage{color}
\usepackage{pgfplots}
\pgfplotsset{compat=1.13}
\usepackage{tikz}
\usepackage{tikz-qtree}
\usepackage{pgfplots}
\usetikzlibrary{patterns}
\DeclareMathOperator*{\argmax}{arg\,max}

\usepackage{amsthm}

\newtheorem{theorem}{Theorem}
\usepackage{footmisc}
\usepackage{url}
\usepackage{cite}
\ifCLASSINFOpdf
\else
\fi

\begin{document}

\title{Minimalistic Attacks: How Little it Takes to Fool Deep Reinforcement Learning Policies}

\author{Xinghua~Qu,~\IEEEmembership{Student Member,~IEEE}, Zhu~Sun,Yew Soon~Ong,~\IEEEmembership{Fellow,~IEEE},
Abhishek~Gupta, Pengfei~Wei 
\thanks{X. Qu, is with Computational Intelligence Lab, School of Computer Science and Engineering, Nanyang Technological University, Singapore, 639798 (email: xinghua001@e.ntu.edu.sg)}
\thanks{Z. Sun is with School of Electrical and Electronic Engineering, Nanyang Technological University, Singapore 639798 (email:sunzhuntu@gmail.com)} 
\thanks{ Y. S. Ong is Director of the Data Science and Artificial Intelligence Research Centre, School of Computer Science and Engineering, Nanyang Technological University, Singapore 639798 (email: asysong@ntu.edu.sg) and Chief Artificial Intelligence Scientist of Singapore's Agency for Science, Technology and Research (email: Ong\_Yew\_Soon@hq.a-star.edu.sg)}
\thanks{A. Gupta is with Singapore Institute of Manufacturing Technology (SIMTech),
A*STAR, Singapore 138634 (email: abhishek\_gupta@simtech.a-star.edu.sg)}
\thanks{P. Wei is with Department of Computer Science, National University of Singapore, Singapore 119077 (email:wpf89928@gmail.com)}}

\markboth{IEEE Transactions on Cognitive and Developmental System}%
{Qu \MakeLowercase{\textit{et al.}}: }

\maketitle

\begin{abstract}
Recent studies have revealed that neural network based policies can be easily fooled by adversarial examples. However, while most prior works analyze the effects of perturbing every pixel of every frame assuming white-box policy access, in this paper we take a more restrictive view towards adversary generation - with the goal of unveiling the limits of a model's vulnerability. 
In particular, we explore minimalistic attacks by defining \textit{\textbf{three key settings}}: (1) \textit{black-box policy access}: where the attacker only has access to the input (state) and output (action probability) of an RL policy;
(2) \textit{fractional-state adversary}: where only several pixels are perturbed, with the extreme case being a single-pixel adversary; and
(3) \textit{tactically-chanced attack}: where only significant frames are tactically chosen to be attacked. 
We formulate the adversarial attack by accommodating the three key settings, and explore their potency on six Atari games by examining four fully trained state-of-the-art policies. In Breakout, for example, we surprisingly find that: (i) all policies showcase significant performance degradation by merely modifying $0.01\%$ of the input state, and (ii) the policy trained by DQN is totally deceived by perturbing only {$1\%$ frames}.
\end{abstract}


\section{Introduction}
Deep learning \cite{mnih2013playing} has been widely regarded as a promising technique in 
reinforcement learning (RL), where the goal of an RL agent is to maximize its expected accumulated reward by interacting with a given environment. 
Although deep neural network (DNN) policies have achieved noteworthy performance on various challenging tasks (e.g., video games, robotics and classical control \cite{li2017adaptive}),
recent studies have shown that these policies are easily deceived under adversarial attacks \cite{huang2017adversarial,lin2017tactics,yuan2019adversarial}. Nevertheless, existing works towards adversary generation are found to make some common simplifying assumptions, namely (1) white-box policy access: where the adversarial examples are computed by back-propagating targeted changes in the prescribed action distribution through known neural network architectures and weights, (2) full-state adversary: where the adversary may perturb almost all pixels of an image representing an input state, and (3) fully-chanced attack: where the attacker strikes the policy at every time step (e.g., at every frame of a video game).

Given that most prior works analyze the effects of perturbing every pixel of every frame assuming white-box policy access, in this paper we propose to take a more restrained view towards adversary generation - with the goal of exploring the limits of a DNN model's vulnerability in RL. We thus focus on \emph{minimalistic attacks} by only considering adversarial examples that perturb limited number of pixels in selected frames, and under restricted black-box policy access. In other words, we intend to unveil how little it really takes to successfully fool state-of-the-art RL policies. In what follows, we briefly introduce the three restrictive settings that characterize our study.

\smallskip\noindent\textbf{Black-box Policy Access (BPA).}
Most previous studies focus on a white-box setting \cite{yuan2019adversarial},
that allows full access to a policy network for back-propagation. However, most systems do not release their internal
configurations (i.e., network structure and weights), only allowing the model to be queried; this makes the white-box assumption too optimistic from an attacker's perspective \cite{chen2017zoo}.
In contrast, we use a BPA setting where the attacker treats the policy as an oracle with access only to its input and outputs.

\begin{figure}[t]
\centering
\begin{subfigure}{0.1\textwidth}
    \centering
     \includegraphics[height=2cm]{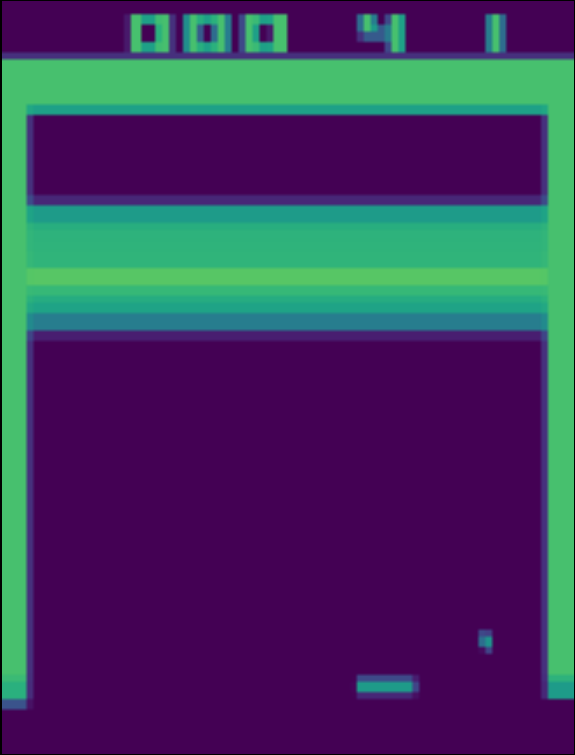}
     \vspace{-0.07in}
     \caption*{\footnotesize Original state}
\end{subfigure}
\begin{subfigure}{0.22\textwidth}
    \centering
     \includegraphics[width=1\linewidth]{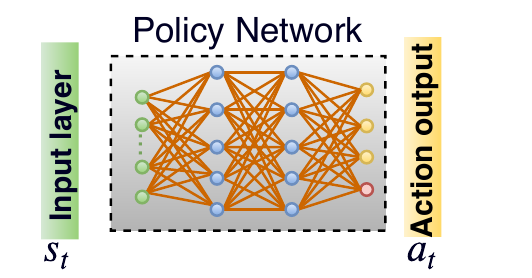}
\end{subfigure}
\hspace{-0.2in}
\begin{subfigure}{0.13\textwidth}
    \centering
     \includegraphics[width=1\linewidth]{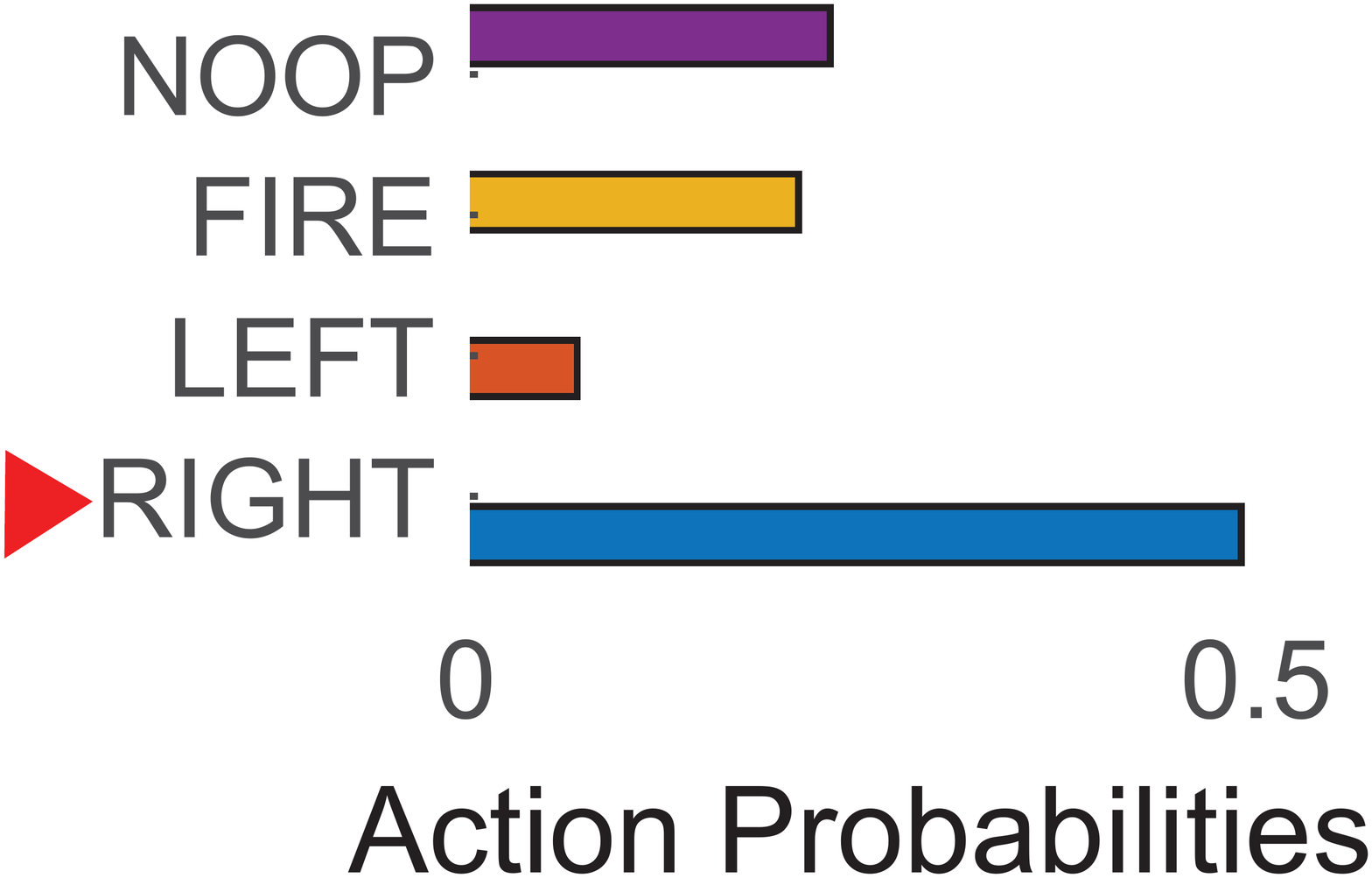}
\end{subfigure}\\
\vspace{0.1in}
\begin{subfigure}{0.1\textwidth}
    \centering
     \includegraphics[height=2cm]{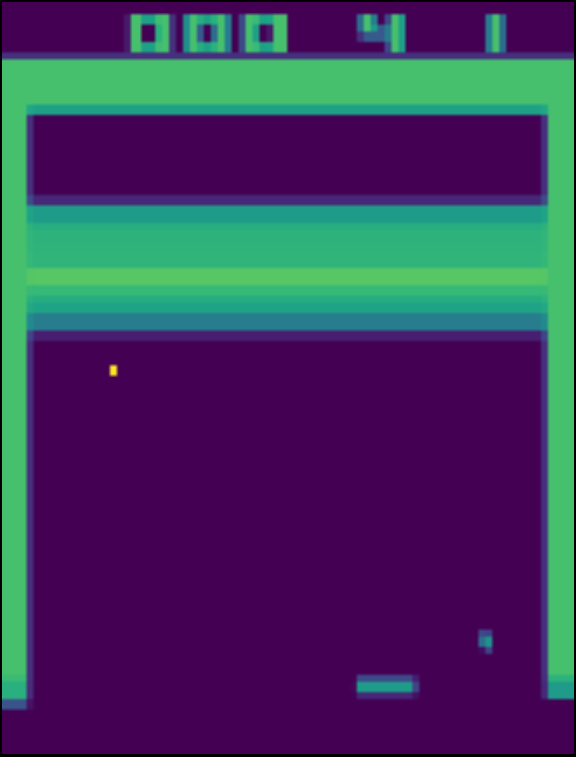}
     \vspace{-0.07in}
     \caption*{\footnotesize Perturbed state}
\end{subfigure}
\begin{subfigure}{0.22\textwidth}
    \centering
     \includegraphics[width=1\linewidth]{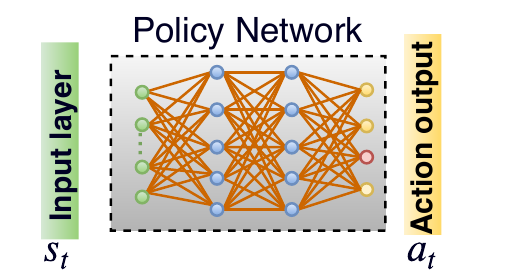}
\end{subfigure}
\hspace{-0.2in}
\begin{subfigure}{0.13\textwidth}
    \centering
     \includegraphics[width=1\linewidth]{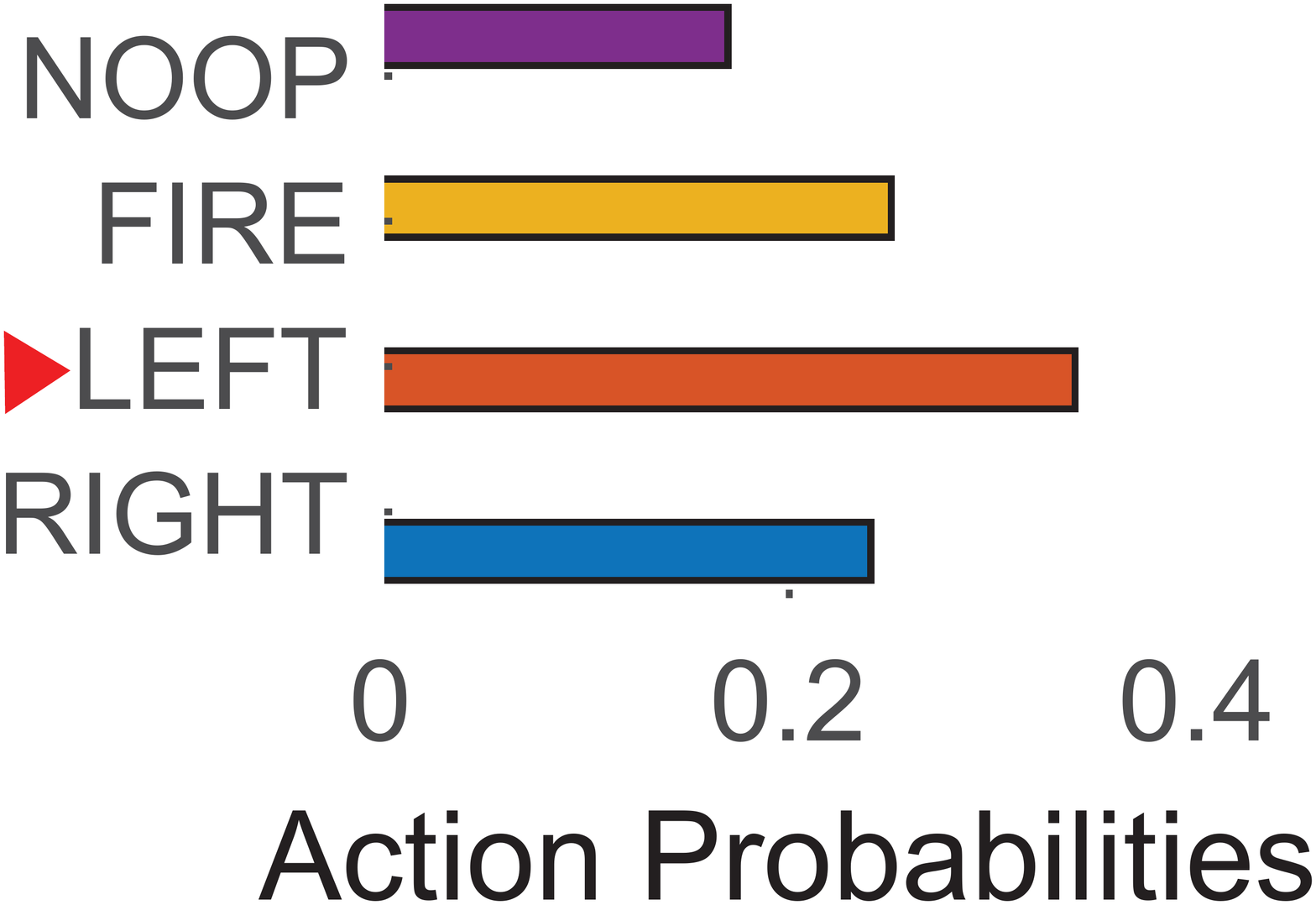}
\end{subfigure}
\vspace{-0.07in}
\caption{The visualization of the single pixel attack on Breakout.}
\label{fig:breakout comparison}
\vspace{-0.2in}
\end{figure}

\smallskip\noindent\textbf{Fractional-State Adversary (FSA).}
In the FSA setting, the adversary only perturbs a small fraction of the input state. This, in the extreme situation, corresponds to the single-pixel attack shown in Fig. \ref{fig:breakout comparison}, where perturbing a single pixel of the input state is found to change the action prescription from `RIGHT' to `LEFT'. In contrast, most previous efforts \cite{yuan2019adversarial}
are mainly based on a full-state adversary (i.e., the number of modified pixels is fairly large, usually spanning the entire frame).

\smallskip\noindent\textbf{Tactically-Chanced Attack (TCA).}
In previously studied RL adversarial attacks \cite{huang2017adversarial,pattanaik2018robust,pinto2017robust}, the adversary strikes the policy on every frame of an episode; this is a setting termed as the fully-chanced attack. Contrarily, we investigate a relatively restrictive case where the attacker only strikes at a few selected frames - a setting we term as \textit{tactically-chanced attack}, where a minimal number of frames is explored to strategically deceive the policy network.

The proposed restrictive settings are deemed to be significant for safety-critical RL applications, such as in the medical treatment of sepsis patients in intensive care units \cite{raghu2019reinforcement,raghu2017continuous} and treatment strategies for HIV \cite{parbhoo2017combining}. 
In such domains, there may exist a temporal gap between the acquisition of the input and the action execution; thereby, providing a time window in which a tactical attack could take place.
In the future, as we move towards integrating end-to-end vision-based systems for healthcare, the sensitivity of prescribed medical actions (such as drug prescription or other medical interventions) to perturbations in medical images poses a severe threat to the utility of RL in such domains. Moreover, the restrictive nature of the attacks makes them practically imperceptible, greatly reducing the chance of identifying and rectifying them. With this in mind, the present paper explores the vulnerability of deep RL models to restrictive adversarial attacks, with particular emphasis on image-based tasks (i.e., Atari games).

The key challenge of this research, therefore, lies in assessing how to effectively accommodate the three restrictive settings for adversarial attack generation in deep RL.
To this end, we design a mathematical program with a novel objective function to generate the FSA under black-box settings, and propose an entropy-based uncertainty measurement to achieve the TCA. 
The optimization variables are defined as the discrete 2-D coordinate location(s) and perturbation value(s) of the selectively attacked pixel(s), and the designed mathematical program guarantees a successful deception of the policy as long as a positive objective value is found. The optimization procedure, for a given input frame, is then carried out by a simple genetic algorithm (GA) \cite{holland1992genetic,deep2009real}.
Furthermore, the Shannon entropy of the action distribution is specified as the attack uncertainty, and is utilized to only select a few salient frames to be tactically attacked. 
We demonstrate the sufficiency of the three restrictive settings on four state-of-the-art deep RL policies pre-trained on six Atari games. 

To sum up, the contributions of this paper can be encapsulated as follows:
\begin{itemize}
    \item We unveil how little  it takes to deceive an RL policy by considering three restrictive settings, namely, black-box policy access, fractional-state adversary, and tactically-chanced attack; 
    \item We formulate the RL adversarial attack as a black-box optimization problem comprising a novel objective function and discrete FSA optimization variables. We also propose a Shannon entropy-based uncertainty measurement to sparingly select the most vulnerable frames to be attacked;
    \item  We explore the three restrictive settings on the policies trained by four state-of-the-art RL algorithms (i.e., DQN \cite{mnih2013playing}, PPO \cite{schulman2017proximal}, A2C \cite{mnih2016asynchronous}, ACKTR \cite{wu2017scalable}) on six representative Atari games (i.e., Pong, Breakout, SpaceInvaders, Seaquest, Qbert, BeamRider);
    \item  Surprisingly, we find that on Breakout: (i) with only \textbf{\textit{a single pixel}} ($\approx0.01\%$ of the state) attacked, the trained policies show significant performance degradation; and (ii) by merely attacking around {$1\%$ frames}, the policy trained by DQN is significantly deceived.
\end{itemize}

The remainder of this paper is organized as follows. In Section II, we provide an overview of related works in RL adversarial attacks, highlighting the novelty of our paper. Section III presents preliminaries on reinforcement learning and adversarial attacks. In section IV, we describe our proposed optimization problem formulation and the associated procedure to generate adversarial attacks for deep RL policies. This is
followed by Section V, where
we report numerical results demonstrating the effectiveness of our attacks.
Finally, Section VI concludes this paper.

\section{Related Work}
Since Szegedy et al. \cite{szegedy2013intriguing}, a number of adversarial attack methods have been investigated to fool deep neural networks (DNNs). Most existing approaches generate adversarial examples through perturbations to images that are imperceptible to the human eye \cite{goodfellow2014explaining,kurakin2016adversarial,xiao2019characterizing}, with the goal of deceiving trained classifiers.
As such, we find that in the recent past, attention has largely been focused on adversarial attacks towards supervised learning tasks. In contrast, adversarial attacks in RL have been relatively less explored to date.   

In RL, Huang et al. \cite{huang2017adversarial} were among the first to demonstrate that neural network policies are vulnerable to adversarial attacks by adding small modifications to the input state of Atari games. 
A full-state adversary (in which an adversarial example may perturb every pixel in the input state) has previously been generated by white-box policy access \cite{goodfellow2014explaining}, where adversarial examples are computed by back-propagating targeted changes to the action distribution prescribed by the policy.
Lin~et~al.~\cite{lin2017tactics} proposed strategically-timed attacks and the so-called enchanting attack, but the adversary generation is still based on a white-box policy access assumption and full-state perturbation. Apart from the above, Kos et al. \cite{kos2017delving} compared the influence of full-state perturbations with random noise, and utilized the value function to guide the adversary injection. 

In summary, existing works are largely based on white-box policy access, together with assumptions of a full-state adversary and fully-chanced attack. There is little research studying the potency of input perturbations that may be far less extensive. Therefore, our goal in this paper is to investigate a significantly more restrained view towards analyzing the vulnerability of deep RL models, viz., based on black-box policy access, fractional-state adversary, and tactically-chanced attacks. Additionally,it is contended that studying such restrictive scenarios might give new insights on the geometrical characteristics and overall behavior of deep neural networks in high dimensional spaces \cite{fawzi2017robustness}.

\begin{figure}[t]
    \centering
    \includegraphics[width=0.8\linewidth]{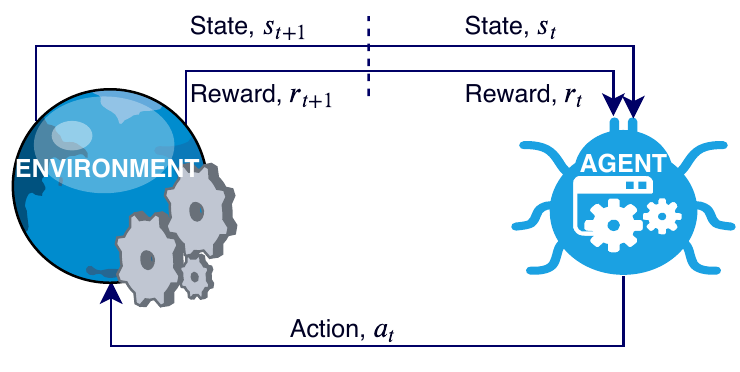}
    \caption{Markov Decision Process in Reinforcement Learning: according to the state $s_t$, the agent selects an action $a_t$, and then receives the reward $r_{t+1}$ from the environment.
    }
    \label{fig:mdp}
    \vspace{-0.2in}
\end{figure}

\section{Preliminaries}
This section first provides some background on RL and several representative approaches to learn RL policies. It then illustrates the basic concepts of adversarial attacks.

\subsection{Reinforcement Learning}
Reinforcement learning (RL) \cite{qu2019memetic,xie2018extended,de2018learning,hou2017evolutionary,zhang2019collaborative} can be formulated as a Markov Decision Process (MDP) \cite{papadimitriou1987complexity}, where the agent interacts with an environment based on the reward  and state transition. This decision making process is shown in Fig. \ref{fig:mdp},
where $s_{t}$ and $r_t$ are the state and reward received from environment at step $t$, and $a_{t}$ is the action selected by the agent. Based on $a_{t}$, the agent interacts with the environment, transitioning to state $s_{t+1}$ and receiving a new reward $r_{t+1}$. This procedure continues until the end of the MDP.

The aim of RL \cite{schulman2015trust} is to find an optimal policy $\pi(\theta^*)$ that maximizes the expected accumulated reward,
\begin{equation}
    \theta^* = \max_{\theta}\mathbb{E}[\sum\nolimits_{t=0}^{T}\gamma^t r_t|\pi_{\theta}],
\end{equation}
where $\theta$ represents the policy parameters (e.g., the weights of a neural network); $\sum_{t=0}^{T}\gamma^t r_t$ is the sum of discounted rewards over an episode; and $\gamma\in(0,1)$
denotes the discount factor that balances the long- and short-term rewards. 

To find the optimal policy parameters $\theta^*$, many different RL algorithms have been proposed. We select several representative ones for demonstration, including DQN \cite{mnih2013playing}, PPO \cite{schulman2017proximal}, A2C \cite{mnih2016asynchronous}, ACKTR \cite{wu2017scalable}.
\begin{itemize}
\item \textbf{Deep Q-Networks (DQN) \cite{mnih2013playing}}: Instead of predicting the probability of each action, DQN computes the \emph{Q value} for each available action. Such Q value represents the approximated accumulated reward from the current frame. Based on such approximation, DQN is trained via minimizing the temporal difference loss (i.e., squared Bellman error). In this paper, the corresponding policy for DQN is achieved by selecting the action with maximum Q value. To keep the DQN-trained policy consistent with the other three algorithms, a soft-max layer is added at the end of the network. 

\item \textbf{Proximal Policy Optimization (PPO)\cite{schulman2017proximal}}: PPO is an off-policy  method using policy gradient, and it pursues
a balance among ease of implementation, sample complexity, and ease of tuning. PPO is proposed based on trust region policy optimization (TRPO) that solves a constrained optimization problem so as to alleviate performance instability. However, PPO handles this issue in a different manner as compared to TRPO; it involves a penalty term that indicates the Kullback–Leibler divergence between the old policy action prediction and the new one. This operation computes an update at each step, minimizing the cost function while restricting the updating step to be relatively small. 

\item \textbf{Advantage Actor-Critic (A2C) \cite{sutton2018reinforcement}}: A2C is an actor-critic method that learns both a policy and a state value function (i.e., also called critic). In A2C, the advantage $A$ represents the extra reward if the action $a_t$ is taken, and is formulated as $A = Q(s_t,a_t)-V(s_t)$, where $Q(s_t,a_t)$ is the state-action value and $V(s_t)$ is the state value.
This formulation reduces variance of the policy gradient and thus increases stability of the gradient estimation.  
\item \textbf{Actor Critic using
Kronecker-Factored Trust Region (ACKTR) \cite{wu2017scalable}}: ACKTR is an extension of the natural policy gradient, which optimizes both the actor and the critic using Kronecker-factored approximate curvature (K-FAC) with trust region. To the best of knowledge, ACKTR is the first scalable trust region natural gradient method for actor-critic methods. The efficacy of ACKTR suggests that Kronecker-factored natural gradient approximations in RL is a promising framework.
\end{itemize}

Although notable performances have been achieved by these algorithms on many challenging tasks (e.g., video games \cite{mnih2013playing} and board games \cite{silver2016mastering,silver2017mastering}),
recent studies have revealed that the policies trained by these algorithms are easily fooled by adversarial perturbations \cite{huang2017adversarial,lin2017tactics,yuan2019adversarial}, as introduced next.

\subsection{Adversarial Attack}
Recent studies have shown that deep learning is vulnerable against well-designed input perturbations (i.e., adversaries) \cite{chen2017zoo}. These adversaries can easily fool even seemingly high performing deep learning models through small input perturbations that are often imperceptible to the human eye. Such vulnerabilities have been well studied in supervised learning, and also to a lesser extent in RL \cite{xiao2019characterizing,huang2017adversarial,lin2017tactics}.

In RL, the aim of an adversarial attack is to find the optimal adversary $\delta_t$ that minimizes the accumulated reward. Let $T$ represent the length of an episode, and $r(\pi,s_t)$ indicate a function that returns the reward given state $s_t$ and DNN based policy $\pi$. Accordingly, the problem setting for generating adversarial attacks in RL can be formulated as,
\begin{equation}\small
    \min_{\delta_t} \sum\nolimits_{t=1}^{T} r(\pi,s_t+\delta_t)
   : \forall t \norm{\delta_t} \leqslant {L}
    \label{equ:adversarialattack problem}
\end{equation}
where the adversary generated at time-step $t$ is represented by $\delta_t$, and its norm (i.e., $\norm{\delta_t}$) is bounded by ${L}$. The basic assumption of 
Eq. (\ref{equ:adversarialattack problem}) is that the misguided action selection of policy $\pi$ will result in a reward degradation. In other words, an action prediction $a_t^p$ obtained from the perturbed state $s_t+\delta_t$ may differ from the originally unperturbed action $a_t^o$, thus threatening the reward value $r_t$. This corresponds to the definition of untargeted attacks \cite{chen2017zoo}, which is stated as $a_t^o\neq a_t^p$. 

To solve the optimization problem formulated in Eq. (\ref{equ:adversarialattack problem}), many white-box approaches (e.g., fast gradient sign method \cite{goodfellow2017attacking}) have been applied in previous works \cite{huang2017adversarial,lin2017tactics}. These white-box approaches are essentially derivative-based, as they generate adversarial examples by back-propagating through the policy network to calculate the gradient of a cost function with respect to the input state, i.e., $\nabla_{s_t} J(\pi, \theta,\Delta a_t, s_t)$. Here, $\theta$ represents the weights of the neural network; $\Delta a_t$ is the change in action space; $J$ indicates the loss function (e.g., cross-entropy loss). 

However, an essential precondition of white-box approaches is complete knowledge of the policy, including the model structure and the model parameters $\theta$. Contrarily, a more restrictive setting is black-box policy access~\cite{chen2017zoo}, that only allows an attacker to present the input state to the policy and observe the output. 

In addition to the above,  most prior studies have only analyzed the effects of perturbing every  pixel  of  every  frame in RL, which is deemed too intensive to be of much practical relevance. Therefore, this paper aims to provide a far more restrained view towards adversarial attack generation, with the goal of unveiling how little it takes to fool deep RL policies.

\begin{figure*}[t]
    \centering
     \begin{subfigure}[b]{0.35\textwidth}
         \centering
         \includegraphics[width=\textwidth]{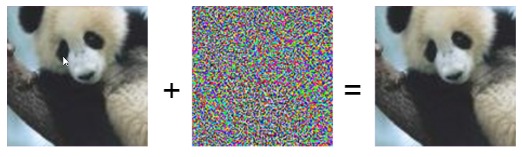}
         \caption*{(a) Full-state adversary \cite{goodfellow2017attacking}}
     \end{subfigure}
     \hspace{0.6in}
     \begin{subfigure}[b]{0.35\textwidth}
         \includegraphics[width=\textwidth]{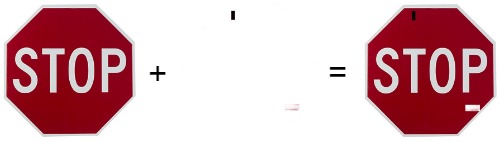}
         \caption*{(b) Fractional-state adversary \cite{brown2017adversarial}}
     \end{subfigure}
     \vspace{-0.05in}
     \caption{Full-state adversary versus the FSA}
     \label{fig:adversary examples}
     \vspace{-0.2in}
\end{figure*}

\section{The Proposed Methodology}
This section provides details of the three key ingredients of the proposed restrictive attack setting, namely \textit{black-box policy access} (BPA), \textit{fractional-state adversary} (FSA), and \textit{tactically-chanced attack} (TCA).

\subsection{Black-box Policy Access}
We adopt the commonly used black-box definition \cite{chen2017zoo,su2019one} from supervised learning, where the attacker can only query the policy. 
In other words, the attacker is unable to compute the gradient $\nabla_{s_t} \pi(\cdot|s_t,\delta_t)$, but only has the privilege to query a targeted policy so as to obtain useful information for crafting adversarial examples. We note that such a setting has been rarely investigated in RL adversarial attacks.  

We realize the proposed BPA setting by formulating the adversarial attack as a black-box optimization problem. 
To this end, we first define a measure $\mathcal{D}(\cdot)$  to quantify the discrepancy between the original action distribution $\pi(\cdot|s_t)$ (produced by an RL policy without input perturbation) and the corresponding perturbed distribution $\pi(\cdot|s_t+\delta_t)$. 
Assuming a finite set of $m$ available actions $a_t^1,a_t^2,\cdots a_t^m$, the probability distribution over actions is represented as $\pi(\cdot|s_t)=[p(a_t^1), p(a_t^2),\cdots, p(a_t^m)]$, where $\sum_{j=1}^m p(a_t^j) = 1$.
Typically, a deterministic policy selects the action $o=\argmax_j p(a_t^j)$.
With this, the black-box attacker considers the discrepancy measure as the optimization objective function, where the state $s_t$ is perturbed by adversary $\delta_t$ such that the measure $\mathcal{D}(\cdot)$ is maximized.

The overall problem formulation can thus be stated as,
\begin{equation}\small
\begin{split}
    \max_{\delta_t}&\text{ }\mathcal{D}(\pi(\cdot|s_t),\pi(\cdot|s_t+\delta_t)): \forall t \norm{\delta_t} \leqslant {L}.
\end{split}
\vspace{-0.1in}
\label{equ:black-box optimization}
\end{equation}
%

The above mathematical program changes the action distribution to $\pi(\cdot|s_t+\delta_t)$, from the original (optimal) action distribution $\pi(\cdot|s_t)$. \textit{If the change in action distribution leads to a change in action selection at any point in the trajectory of an RL agent, then, based on Bellman's principle of optimality~\cite{sutton2018reinforcement}, the resulting sub-trajectory shall lead to a sub-optimal cumulative reward as a consequence of the attack}.
It is based on this principle that we propose to substitute the reward minimization problem of Eq. (\ref{equ:adversarialattack problem}) with the discrepancy maximization formulation in Eq. (\ref{equ:black-box optimization}).

The exact choice of the discrepancy measure $\mathcal{D}(\cdot)$ is expected to have a significant impact on the attack performance, as different measures shall capture different variations in the action distributions. Previous works~\cite{huang2017adversarial,xiao2019characterizing,lin2017tactics} apply the Euclidean norm (e.g., $L_1$, $L_2$, $L_\infty$) between $\pi(\cdot|s_t+\delta_t)$ and $\pi(\cdot|s_t)$. However, such $L_p$ norm cannot guarantee a successful untargeted attack, since maximizing the $L_p$ norm may not guarantee that the action selection has been altered by the perturbed state. Thus, a successful untargeted attack under deterministic action selection must ensure that $ \argmax_j\pi(\cdot|s_t+\delta_t)_j \neq \argmax_j\pi(\cdot|s_t)_j$, where $\pi(\cdot|s_t)_j = p(a_t^j)$.

To enable a more consistent discrepancy measure for untargeted attacks, we design the following function $\widetilde{\mathcal{D}}$ based on the query feedback of the policy,
\begin{equation}
\begin{split}
    \widetilde{\mathcal{D}}=&\max_{j\neq o}\pi(\cdot|s_t+\delta_t)_j -\pi(\cdot|s_t+\delta_t)_{o}\\
    &\text{where: }
     o = \argmax_j \pi(\cdot|s_t)_j.\\
\end{split}
\vspace{-0.1in}
\label{equ:maxmax_distance}
\end{equation}
This formulation is different from the Euclidean norm, guaranteeing a successful untargeted attack if $\widetilde{\mathcal{D}}$ is positive. To support this claim, we refer to the theorem and proof below.
\begin{theorem} Suppose the discrepancy measure $\widetilde{\mathcal{D}}$ in Eq. (\ref{equ:maxmax_distance}) is positive and policy $\pi$ is a deterministic, i.e., action $a_t^o$ is chosen such that $o = \argmax_j\pi(\cdot|s_t)$. Then, the adversarial example $\delta_t$ for policy $\pi$ at state $s_t$ is guaranteed to be a successful untargeted attack, i.e.,
\begin{equation*}
    \argmax_{j}[\pi(\cdot|s_t+\delta_t)]_j\neq \argmax_{j}[\pi(\cdot|s_t)]_j.
\end{equation*}
\end{theorem}
\vspace{-0.1in}
\begin{proof}
We use the symbol $o$ and $p$ to represent the index of the selection action from state $s_t$ and perturbed state $s_t+\delta_t$, respectively. As the policy $\pi$ is deterministic, $o$ and $p$ are given as,
\begin{equation*}
    \begin{split}
    o &= \argmax_j[\pi(\cdot|s_t)]_j,\\
    p & = \argmax_j[\pi(\cdot|s_t+\delta_t)]_j.\\
\end{split}
\end{equation*}
If the discrepancy measure $ \widetilde{\mathcal{D}}$ is positive, then we have,
\begin{equation*}
\begin{split}
  \widetilde{\mathcal{D}}>0 \Rightarrow&\max_{j\neq o}\pi(\cdot|s_t+\delta_t)_j -\pi(\cdot|s_t+\delta_t)_{o}>0\\
    \Rightarrow& \max_{j\neq o}\pi(\cdot|s_t+\delta_t)_j >\pi(\cdot|s_t+\delta_t)_{o}\\
   \Rightarrow&\max_{j}\pi(\cdot|s_t+\delta_t)_j = \max_{j\neq o}\pi(\cdot|s_t+\delta_t)_j. \\
\end{split}
\end{equation*}
Therefore, the perturbed action index $p$ is given by,  
\begin{equation*}
\begin{split}
p&=\argmax_{j}\pi(\cdot|s_t+\delta_t)_j=\argmax_{j\neq o}\pi(\cdot|s_t+\delta_t)_j\neq o \\
&\Rightarrow\argmax_{j}[\pi(\cdot|s_t+\delta_t)]_j \neq \argmax_{j}[\pi(\cdot|s_t)]_j.\qedhere
\end{split}
\end{equation*}
\vspace{-0.15in}
\end{proof}

With the aforementioned discrepancy measure $\widetilde{\mathcal{D}}$, the mathematical program for generating attacks is given by, 
\begin{equation}
\begin{split}
    \max_{\delta_t}&\text{ }\max_{j\neq o}\pi(\cdot|s_t+\delta_t)_j -\pi(\cdot|s_t+\delta_t)_{o} \\
     &\text{where:  } \forall t\  \norm{\delta_t} \leqslant {L}\text{, } o = \argmax_j\pi(\cdot|s_t)_j
\end{split} 
\label{equ:BPA optimization}
\vspace{-0.1in}
\end{equation}

When resolving the optimization problem in Eq. (\ref{equ:BPA optimization}), $\delta_t$ is determined according to its parameterisation. In this paper, $\delta_t$ is limited to perturbing only a small fraction of the input state (i.e., fractional-state adversary), as described next. 
\subsection{Fractional-State Adversary}

To explore adversarial attacks limited to a few pixels in RL, we investigate the fractional-state adversary (FSA) setting.
In comparison to the full-state adversary depicted in Fig.~\ref{fig:adversary examples}(a), FSA only perturbs a fraction of the state; shown in Fig.~\ref{fig:adversary examples}(b). 
The extreme scenario for FSA is merely a single-pixel attack, which is deemed to be more physically realizable for the attacker than a full-state one. For instance, pasting a sticker or a simple fading of color in the ``STOP" sign could easily lead to a FSA in Fig. \ref{fig:adversary examples}(b).

To achieve the FSA setting, we  parameterize the adversary $\delta_t$ by its corresponding pixel coordinates (i.e., $\textbf{x}_t,\textbf{y}_t$) and perturbation value $\textbf{p}_t$ as follows,
\begin{equation}
\begin{split}
    \delta_t&\leftarrow\mathcal{P}(\textbf{x}_t,\textbf{y}_t,\textbf{p}_t)\\
    &\leftarrow\mathcal{P}(x_{t}^1,y_{t}^1,p_{t}^1, \cdots, x_{t}^n,y_{t}^n,p_{t}^n)\\
\end{split}
\label{equ:delta}
\end{equation}
where $n$ is the number of pixels that are attacked in a frame, ${x}_t^i,{y}_t^i$ are coordinate values of the $i^{th}$ perturbed pixel, and ${p}_t^i$ is the adversarial perturbation value of the $i^{th}$ pixel. With Eq. (\ref{equ:delta}), the final black-box optimization problem is stated as,
\begin{equation}
\small
    \begin{split}
        \max_{\textbf{x}_t,\textbf{y}_t,\textbf{p}_t}&\max_{j\neq o}[\pi(\cdot|s_t+\mathcal{P}(\textbf{x}_t,\textbf{y}_t,\textbf{p}_t))]_j -[\pi(\cdot|s_t+\mathcal{P}(\textbf{x}_t,\textbf{y}_t,\textbf{p}_t))]_{o}\\
     \text{where:  } & \forall t\  o = \argmax_j[\pi(\cdot|s_t)]_j;\\
     & \forall t\  0 \leqslant x_t^i \leqslant \mathcal{I}_\textbf{x}\text{,  }x_t^i\in\mathbb{N}\text{,  }i=[1,2,\cdots,n];\\
     &\forall t\  0 \leqslant y_t^i \leqslant \mathcal{I}_\textbf{y}\text{,  }y_t^i\in\mathbb{N}\text{,  }i=[1,2,\cdots,n];\\
     &\forall t \  0 \leqslant u_t^i+p_t^i \leqslant \mathcal{I}_\textbf{p}\text{,  }p_t^i\in\mathbb{N}\text{,  }i=[1,2,\cdots,n];\\
    \end{split}
    \label{equ:integer programming}
\end{equation}
\begin{algorithm}[t]
\footnotesize
\caption{Attack Generation via GA}\label{GA}
\KwIn{$s_t$ -- state at time-step $t$; $n$ -- number of pixels for attack;
$\lambda$ -- population size; $\beta$ -- selection rate; $\gamma$ -- rate of mutation; $M^{eval}$ -- maximum number of function evaluations; $\zeta^*$ -- TCA threshold value}
Load the policy $\pi$ pre-trained by a particular RL algorithm\;
Obtain the RL input state $s$ in a particular Atari game\;
Calculate the attack uncertainty $\zeta_t$ by Eq. (\ref{equ:attack uncertainty})\;
\uIf{$\zeta_t \geqslant \zeta^*$}
{\tcp{No adversarial attack}
$\delta_t = \texttt{none}$\;}
\Else{
\tcp{Explore the $n$ pixel attack}
Initialize the population $pop$ with randomly generating $\lambda$ individuals\;
\Repeat{the maximum objective evaluations $M^{eval}$ reached}{Evaluate the discrepancy measure $\widetilde{\mathcal{D}}$ of each individual\;
Save the top-$\lambda\beta$ individuals in $pop^\prime$\;
\For{j=1:$\frac{\lambda(1-\beta)}{2}$}{
Crossover: generate $\delta_c^j$ with one-point crossover \cite{holland1992genetic}\;
Mutation: $\delta_m^j = \delta^j+\gamma\cdot\epsilon_j\text{, }\epsilon_j\in \mathcal{N}(0,1)$\;
}
Append $\delta_c$ and $\delta_m$ to $pop'$\;
	    $pop=pop'$}
Return the optimal FSA $\delta^* = [\textbf{x}^*,\textbf{y}^*,\textbf{p}^*]$\;
}
\end{algorithm}\setlength{\textfloatsep}{0.05cm}

\noindent where $\mathcal{I}_\textbf{x}$, $\mathcal{I}_\textbf{y}$ are the integral upper bounds of $\textbf{x}_t$ and $\textbf{y}_t$ respectively, $u_t^i$ is the unperturbed value of the $i^{th}$ pixel in $s_t$, $\mathcal{I}_\textbf{p}$ is the maximum allowable perturbed pixel value, 
and $n$ is the FSA size that controls the number of pixels to be attacked given any input state $s_t$. The extreme case is $n=1$, implying that there is only \textit{one pixel} attacked in an input frame. To the best of our knowledge, such one-pixel attack has never been explored in the context of RL adversarial attacks.

To obtain the optimal $\textbf{x}_t,\textbf{y}_t,\textbf{p}_t$, we utilize a simple genetic algorithm (GA) \cite{holland1992genetic,bertsekas1997nonlinear,deep2009real} (i.e., a \emph{derivative-free} evolutionary computation \cite{jin2005evolutionary,jin2005comprehensive,jin2002framework,jin2018data,liu2005improved,liu2007effective,ma2017tchebycheff,feng2017autoencoding} method for black-box optimization). The pseudo code of the GA is illustrated in Algorithm \ref{GA} (lines 7-17). A population $pop$ containing $\lambda$ individuals is evolved, where each individual $\delta_t^i,i\in[1,\lambda]$ is a candidate adversarial example that is evaluated by the objective function in Eq. (\ref{equ:integer programming}). 
The top-$\lambda\beta$ individuals with respect to the objective value are selected as elites to survive in the new generation (line 10). Based on standard crossover (line 12) and mutation (line 13) operations, another population $pop'$ is reproduced to replace the old $pop$. This process repeats until a specified maximum number of function evaluations $M^{eval}$ is reached. 

\begin{figure*}[ht]
\centering
\begin{subfigure}{0.15\textwidth}
    \centering
     \includegraphics[width=0.8\textwidth]{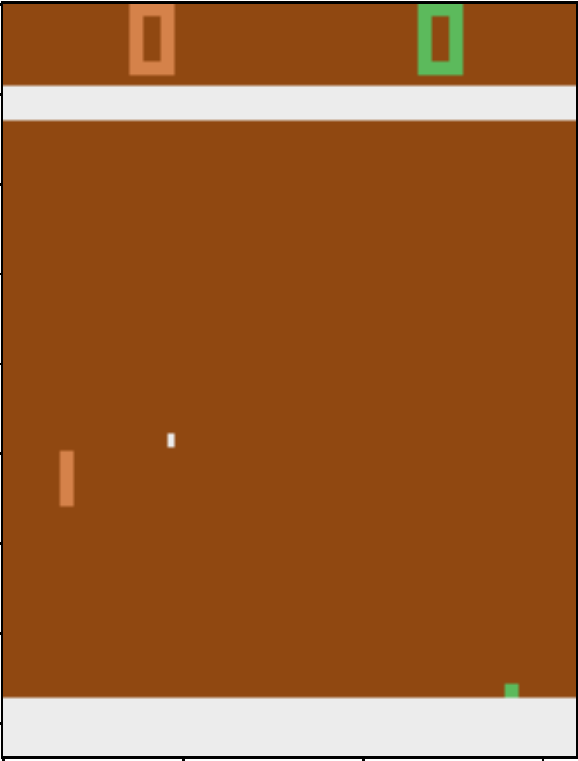}
     \caption{Pong}
\end{subfigure}
\begin{subfigure}{0.15\textwidth}
    \centering
     \includegraphics[width=0.8\textwidth]{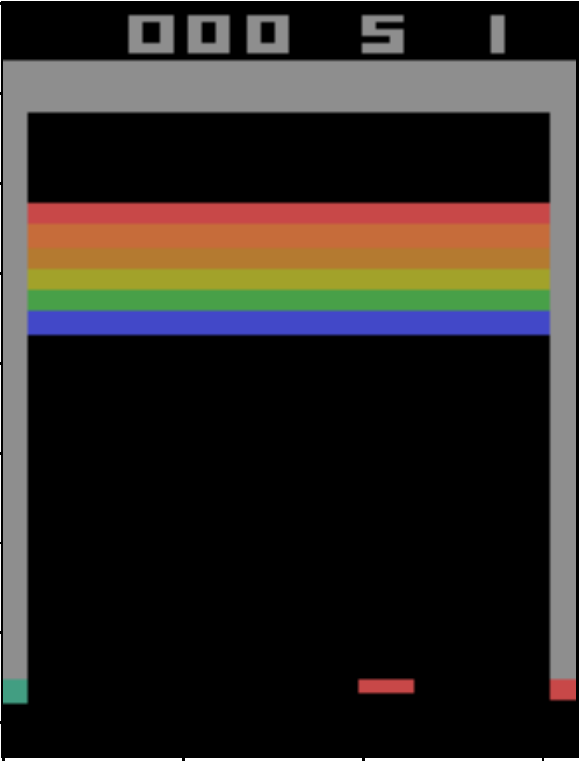}
     \caption{Breakout}
\end{subfigure}
\begin{subfigure}{0.15\textwidth}
    \centering
     \includegraphics[width=0.8\textwidth]{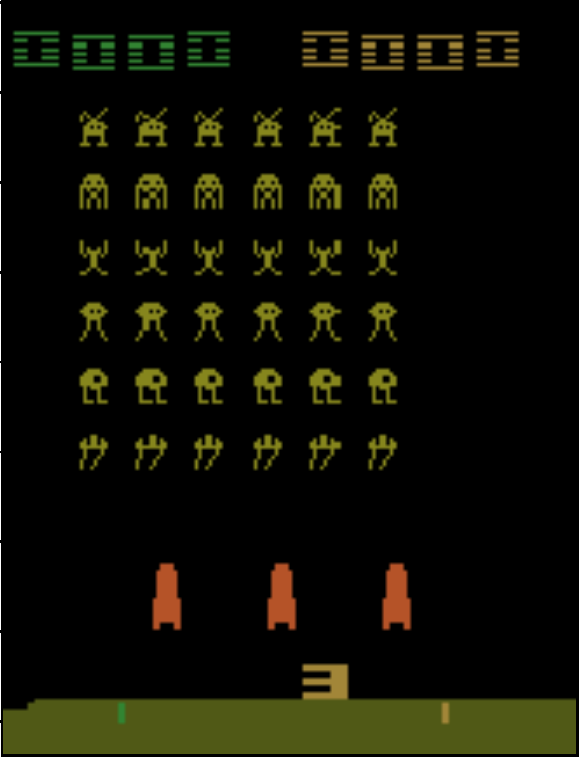}
     \caption{SpaceInvaders}
\end{subfigure}
\begin{subfigure}{0.15\textwidth}
    \centering
     \includegraphics[width=0.8\textwidth]{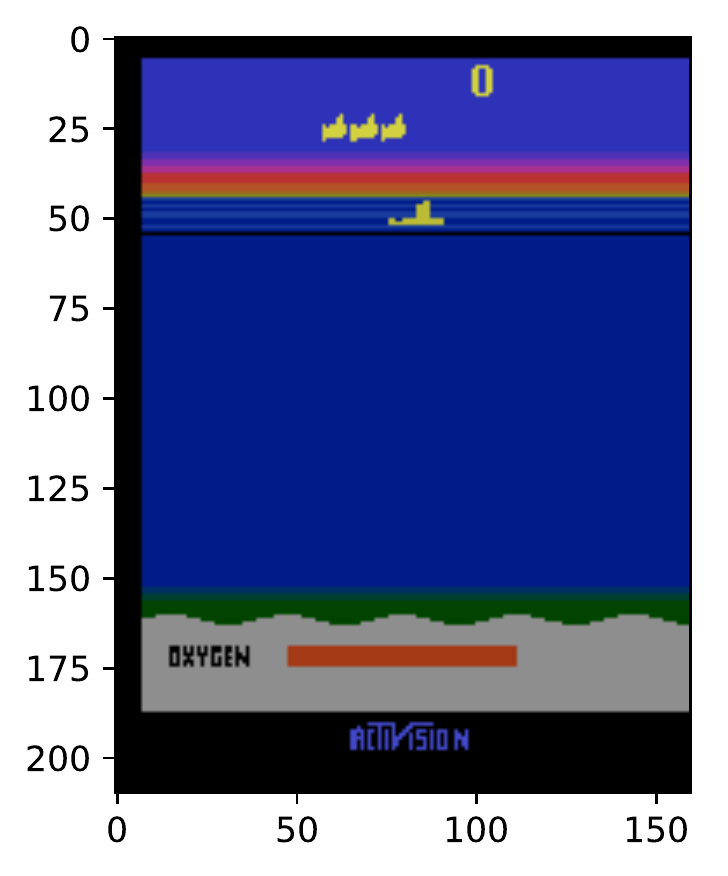}
     \caption{Seaquest}
\end{subfigure}
\hspace{0.02in}
\begin{subfigure}{0.15\textwidth}
    \centering
     \includegraphics[width=0.8\textwidth]{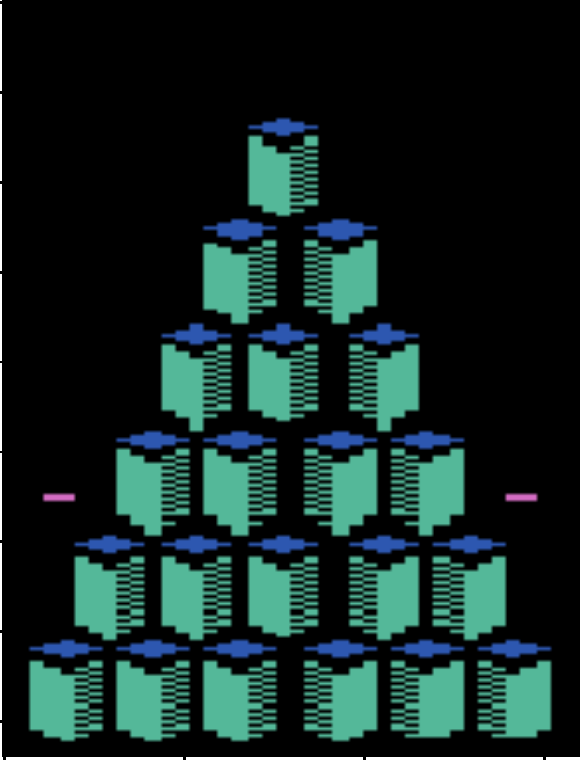}
     \caption{Qbert}
\end{subfigure}
\hspace{0.01in}
\begin{subfigure}{0.15\textwidth}
    \centering
     \includegraphics[width=0.8\textwidth]{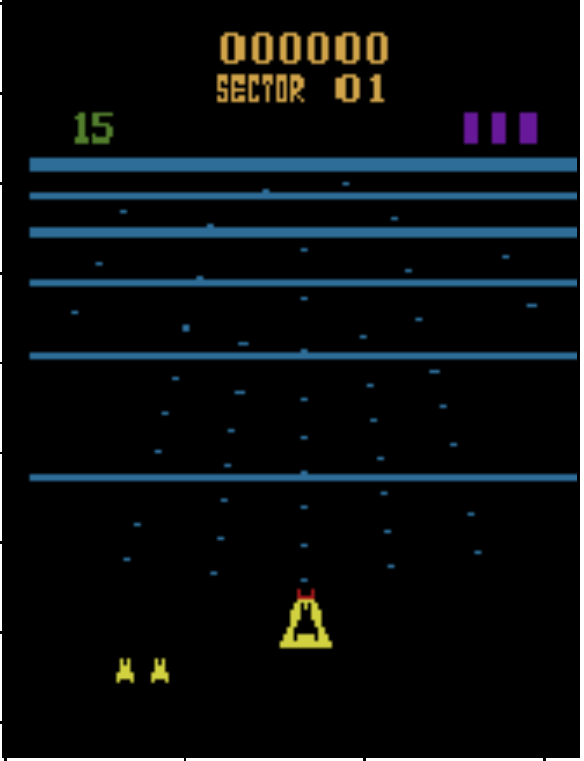}
     \caption{BeamRider}
\end{subfigure}
\caption{The six representative Atari games \cite{brockman2016openai} considered in the experiments}
\label{fig:atari games}
\end{figure*}

\begin{figure}[t]
\centering
\begin{subfigure}{0.4\textwidth}
     \includegraphics[width=\textwidth]{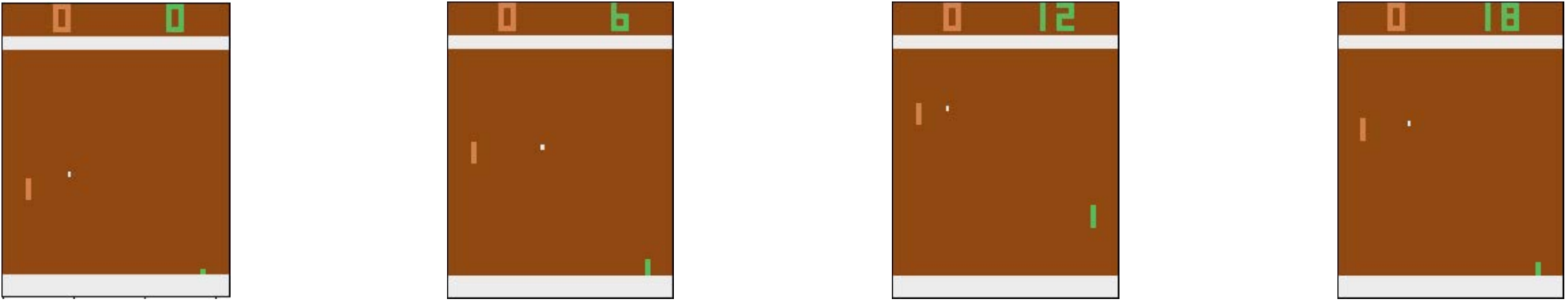}
\end{subfigure}
\begin{subfigure}{0.48\textwidth}
    \centering
     \includegraphics[width=\textwidth]{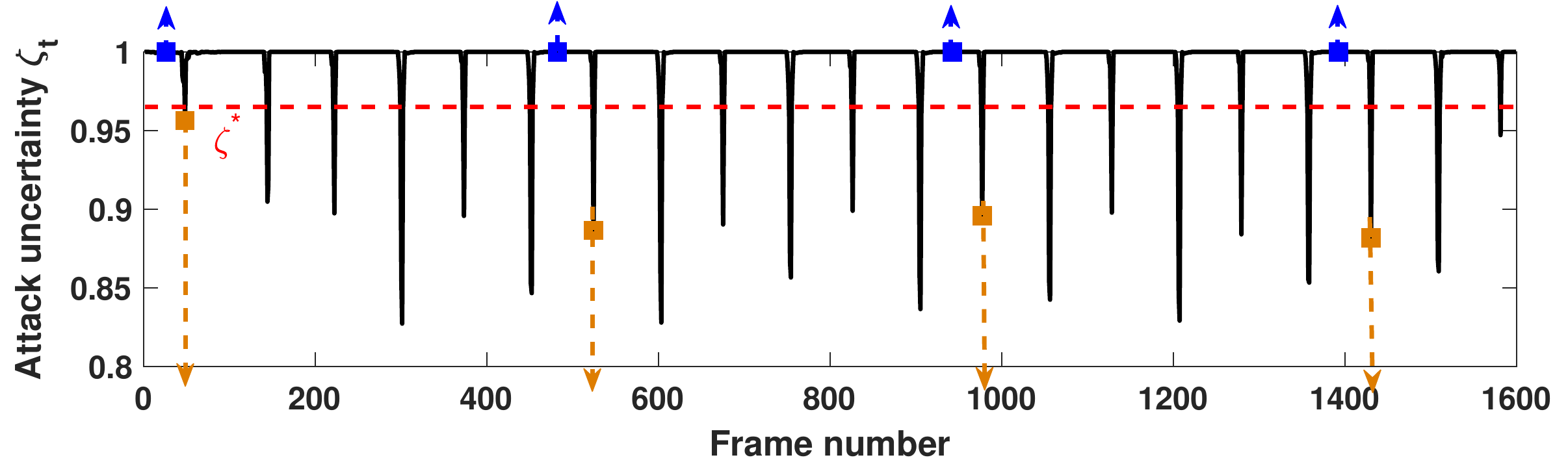}
\end{subfigure}
\begin{subfigure}{0.4\textwidth}
     \includegraphics[width=\textwidth]{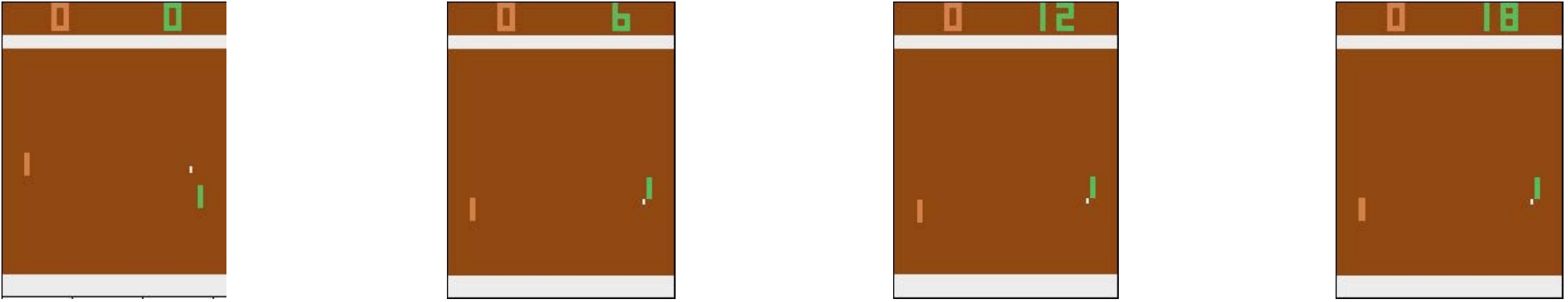}
\end{subfigure}
\caption{The attack uncertainty $\zeta_t$ on Pong.}
\label{fig: Pong entropy}
\end{figure}

\subsection{Tactically-Chanced Attack}
To explore adversarial attacks on a restricted number of frames, we design tactically-chanced attacks (TCA) where the attacker strategically strikes \textit{only} salient frames that are likely to be most contributing to the accumulated reward. In contrast, most existing approaches \cite{huang2017adversarial} apply adversarial examples on every frame, which is referred to as fully-chanced attack. 
Our proposed TCA is more restrictive, realizable, and threatening, as can further be highlighted from three different perspectives: 
(1) due to the communication budget restriction, the attacker may not be able to strike the policy in a fully-chanced fashion;
(2) a tactically-chanced attack is less likely to be detected by the defender; and (3) only striking the salient frames improves the attack efficiency,
as many frames contribute trivially to the accumulated reward. 

To this end, we define the normalized Shannon entropy of the action distribution $\pi(\cdot|s_t) = [p(a_t^1),p(a_t^2),\cdots,p(a_t^m)]$ as a measure of attack uncertainty. Specifically, the attack uncertainty ($\zeta_t$) of each frame is given by,
\begin{equation}
    \zeta_t = -\sum\nolimits_{i=0}^{m}\frac{p(a_t^i)\cdot \log p(a_t^i)}{\log m}
    \label{equ:attack uncertainty}
\end{equation}
where $p(a_i)$ is the probability of action $a_i$; $m$ is the dimensionality of the action space. Note that $\zeta_t$ must lie within $(0,1]$.  
Under this definition, the frame with relatively low $\zeta_t$ indicates that the policy has high confidence in its prescribed action. Hence, we assume that attacking such frames will effectively disrupt the policy.

We demonstrate the attack uncertainty on Pong in Fig. \ref{fig: Pong entropy}; similar trends can be observed on the other games as well.  
In the figure, the frames with smaller $\zeta_t$, marked by brown rectangles, depict that the ball is close to the paddle. On the other hand, the blue rectangle marked ones with larger $\zeta_t$ (close to 1) indicate that the ball is distant from the paddle.
Attacking the brown marked frames are intuitively more effective in fooling the policy, as they are likely to lead to a more considerable reward loss. In contrast, an attack may be relatively inconsequential for the blue marked frames. 
As the ball is distant from the paddle, attacking such frames will have little impact on the cumulative reward.

Therefore, we formulate TCA by defining a TCA threshold ($\zeta^*$), as shown by the red dashed line in Fig. (\ref{fig: Pong entropy}); this threshold controls the proportion of attacked frames, such that only those frames with an uncertainty value below $\zeta^*$ are perturbed.
In the experimental section, we analyze the sensitivity of our approach to the choice of $\zeta^*$ by varying its values; this also helps us to explore how little it actually takes to deceive a policy (from the perspective of the number of frames attacked).
The adversary $\delta_t$ under TCA setting is thus given by,
\begin{equation}\label{equ:adversary_uncertainty}
\small
\delta_t =
\left\{  
\begin{array}{lr}   \mathcal{P}(\textbf{x}_t^*,\textbf{y}_t^*,\textbf{p}_t^*)\text{, if } \zeta_t<\zeta^*,\\
\texttt{none}\text{, if } \zeta_t\geqslant\zeta^*.
\end{array}  
\right.  
\end{equation}

Eq. (\ref{equ:adversary_uncertainty}) implies that if the attack uncertainty ($\zeta_t$) is smaller than $\zeta^*$, an adversary shall be generated by solving the optimization problem in Eq. (\ref{equ:integer programming}). Otherwise, the attacker will tactically hide without wasting compute resources on trivial frames.

\begin{figure}[ht]
    \centering
    \includegraphics[width=0.46\textwidth]{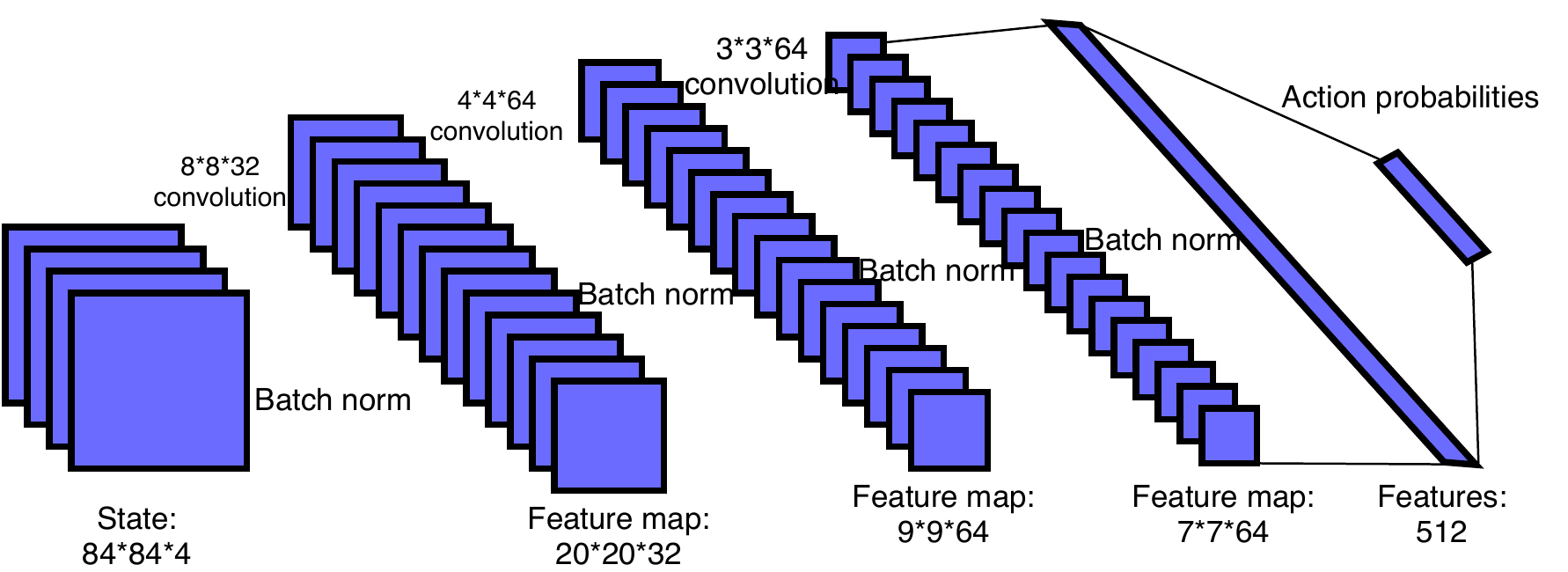}
    \caption{The neural network structure for playing Atari games}
    \label{fig:network_architecture}
\end{figure}


\begin{figure*}[t]
\centering
\includegraphics[width=0.9\textwidth]{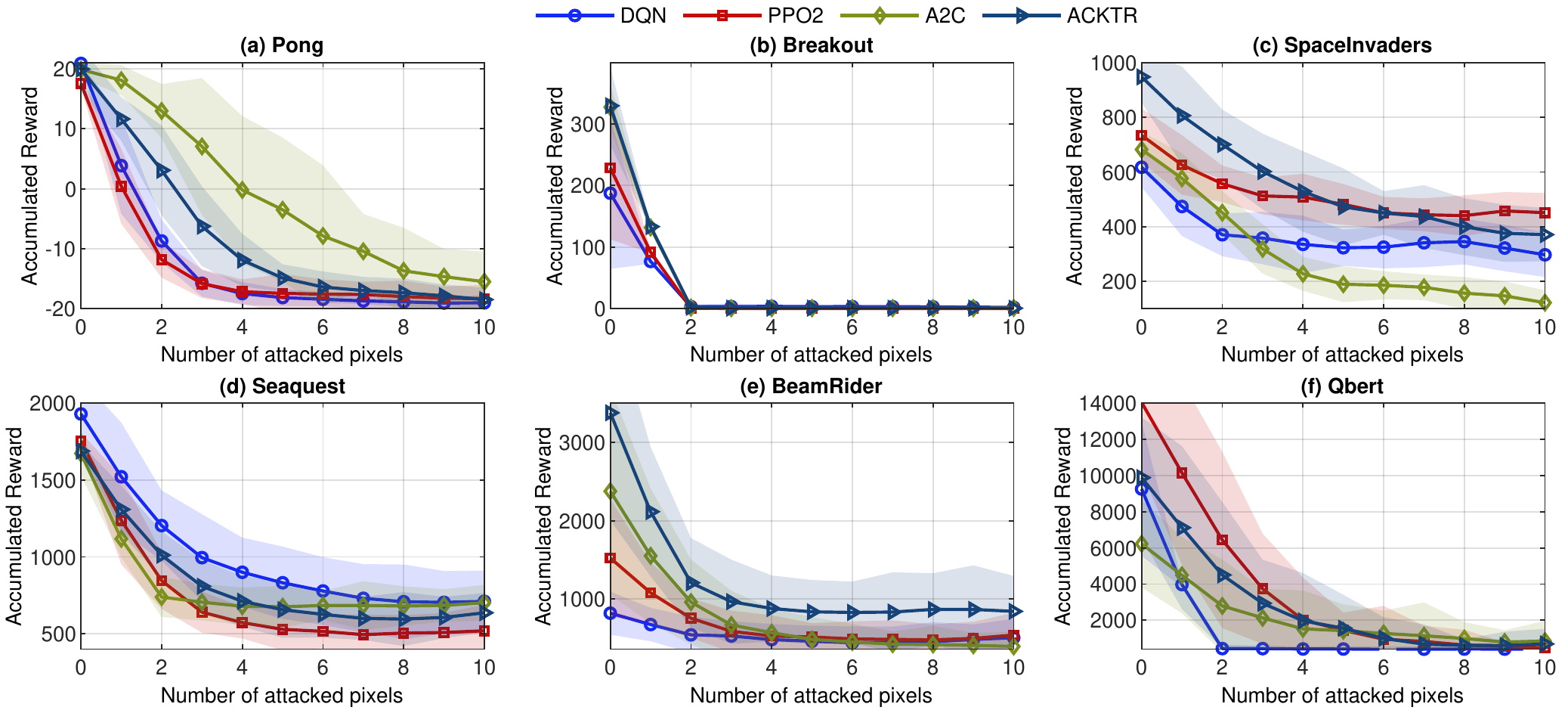}
\caption{The results of adversarial attack with different FSA size $n$ (i.e. number of attacked pixels), where the line and shaded area illustrate the mean and standard deviation of 30 independent runs respectively.}
\label{fig:cubesize_comparison}
\vspace{-0.1in}
\end{figure*}

\begin{figure*}[ht]
\centering
\includegraphics[width=0.9\textwidth]{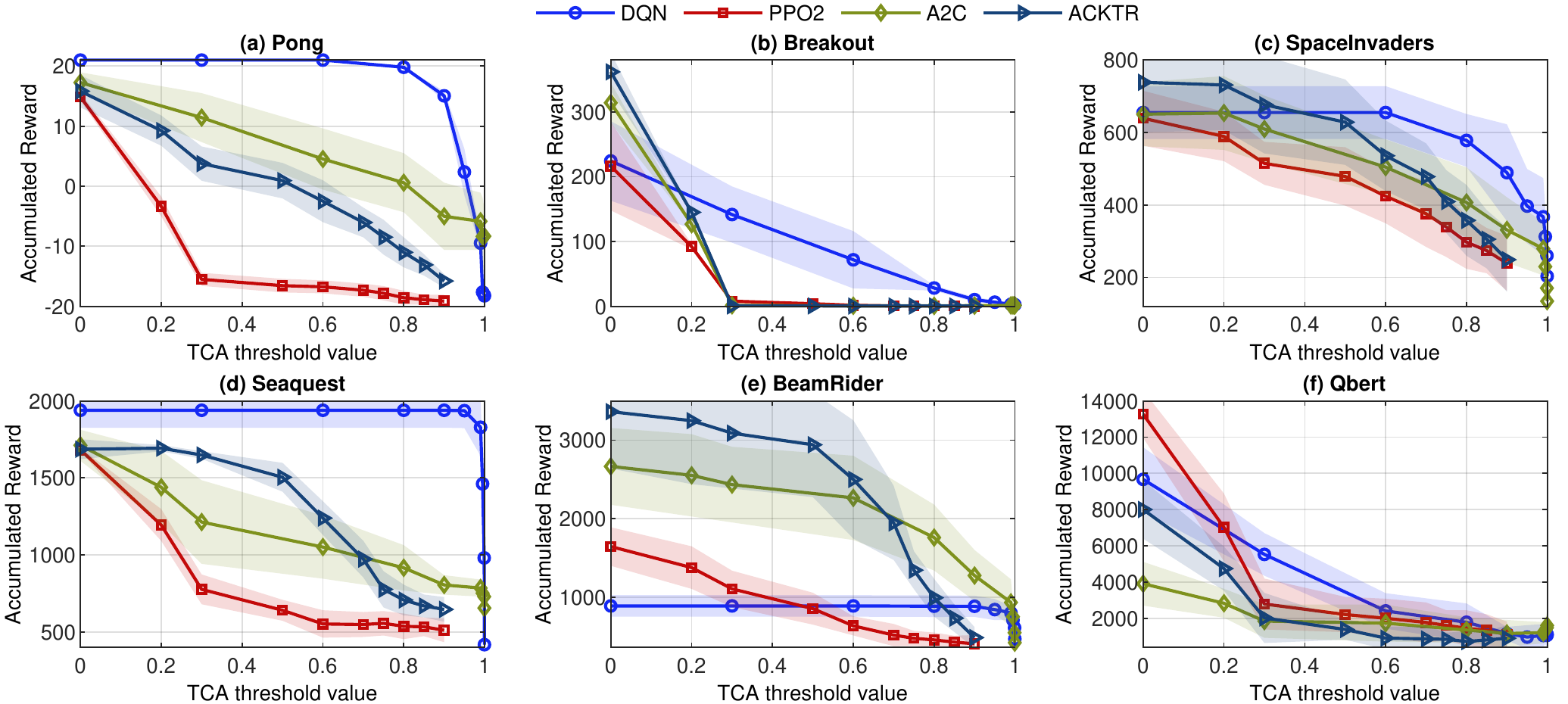}
\caption{The results of adversarial attack  with different TCA threshold $\zeta^*$, where the line and shaded area illustrate the mean and standard deviation of 30 independent runs respectively.}
\label{fig:tca_comparison}
\vspace{-0.2in}
\end{figure*}
\section{Experiments and Analysis}
We evaluate the three restrictive settings on six Atari games in OpenAI Gym \cite{brockman2016openai} with various difficulty levels, including Pong, Breakout, SpaceInvaders, Seaquest, Qbert, and BeamRider\footnote{Our code is available at \footnotesize{\url{https://github.com/RLMA2019/RLMA}}}. These games are shown in Fig.  \ref{fig:atari games}. For each game, the policies are trained by four start-of-the-art RL algorithms, including DQN \cite{mnih2013playing}, PPO2 \cite{schulman2017proximal}, A2C \cite{mnih2016asynchronous}, ACKTR \cite{wu2017scalable}. The network structure is taken from \cite{mnih2015human}, and is kept the same for all the four RL algorithms.

\subsection{Experiment Setup}
We utilize the fully-trained policies from the RL baseline zoo \cite{Araffin2018RLBaselinesZoo}, where each policy is trained for $4\times10^7$ time-steps. The other parameter settings (e.g., learning rate, discount factor and max gradient norm) for each training algorithm follow the default settings in the RL baseline zoo~\cite{Araffin2018RLBaselinesZoo}.
Each policy follows the same pre-processing steps and neural network architecture (i.e., shown as Fig. \ref{fig:network_architecture}) as in \cite{mnih2013playing}. 
The input state $s_t$ of the neural network is the concatenation of the last four screen images, where each image is resized to $84\times84$. 
The pixel value of the grey scale image is in the range $[0,255]$ stepped by 1. 
The output of the policy is a distribution over possible actions for PPO2, A2C, ACKTR, and an estimation of Q values for DQN. 

\begin{figure*}[ht]
\centering
\includegraphics[width=0.9\textwidth]{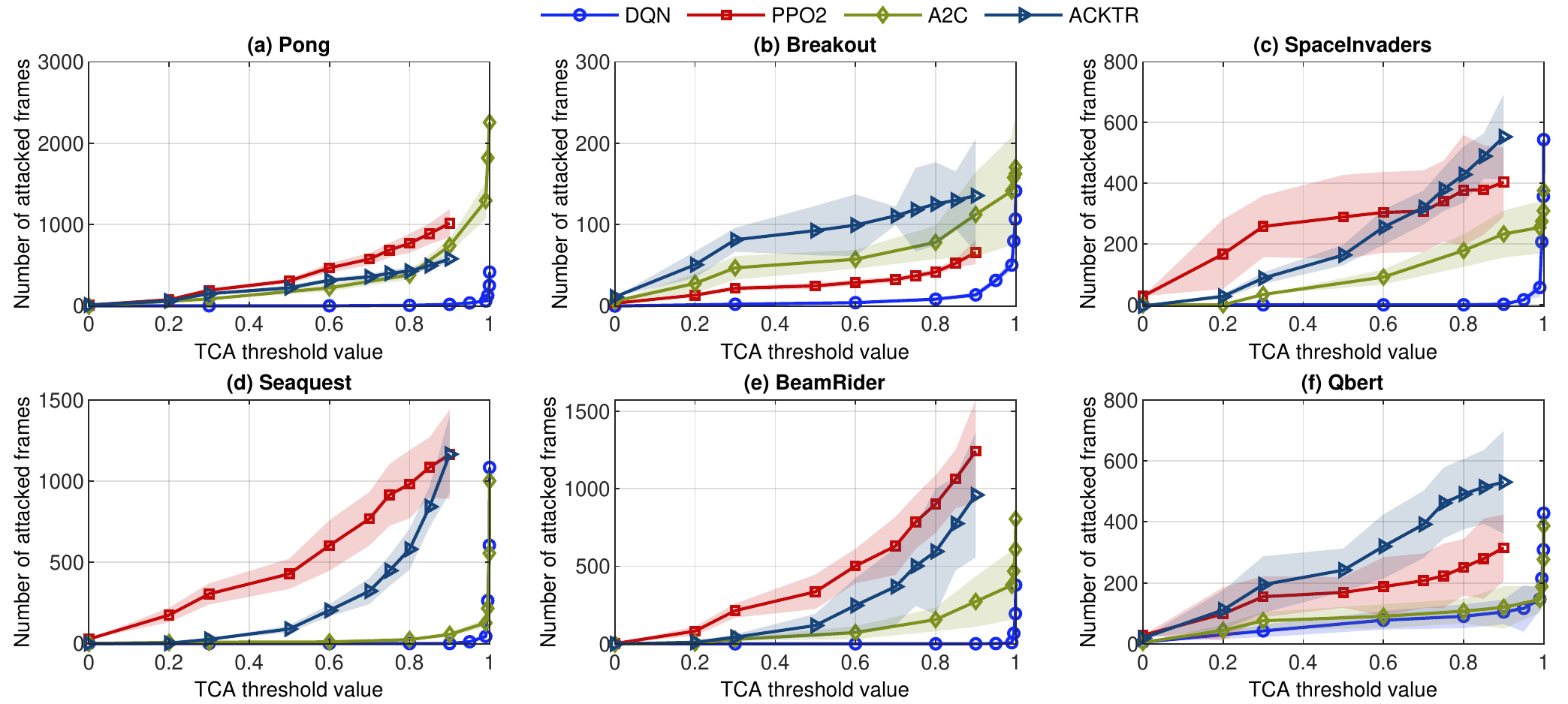}
\caption{The relationship between attacked frames and TCA threshold $\zeta^*$, where the line and shaded area illustrate the mean and standard deviation of 30 independent runs respectively.}
\label{fig:attacked_frames_compare_0}
\end{figure*}

\begin{figure*}[!ht]
\centering
\includegraphics[width=0.9\textwidth]{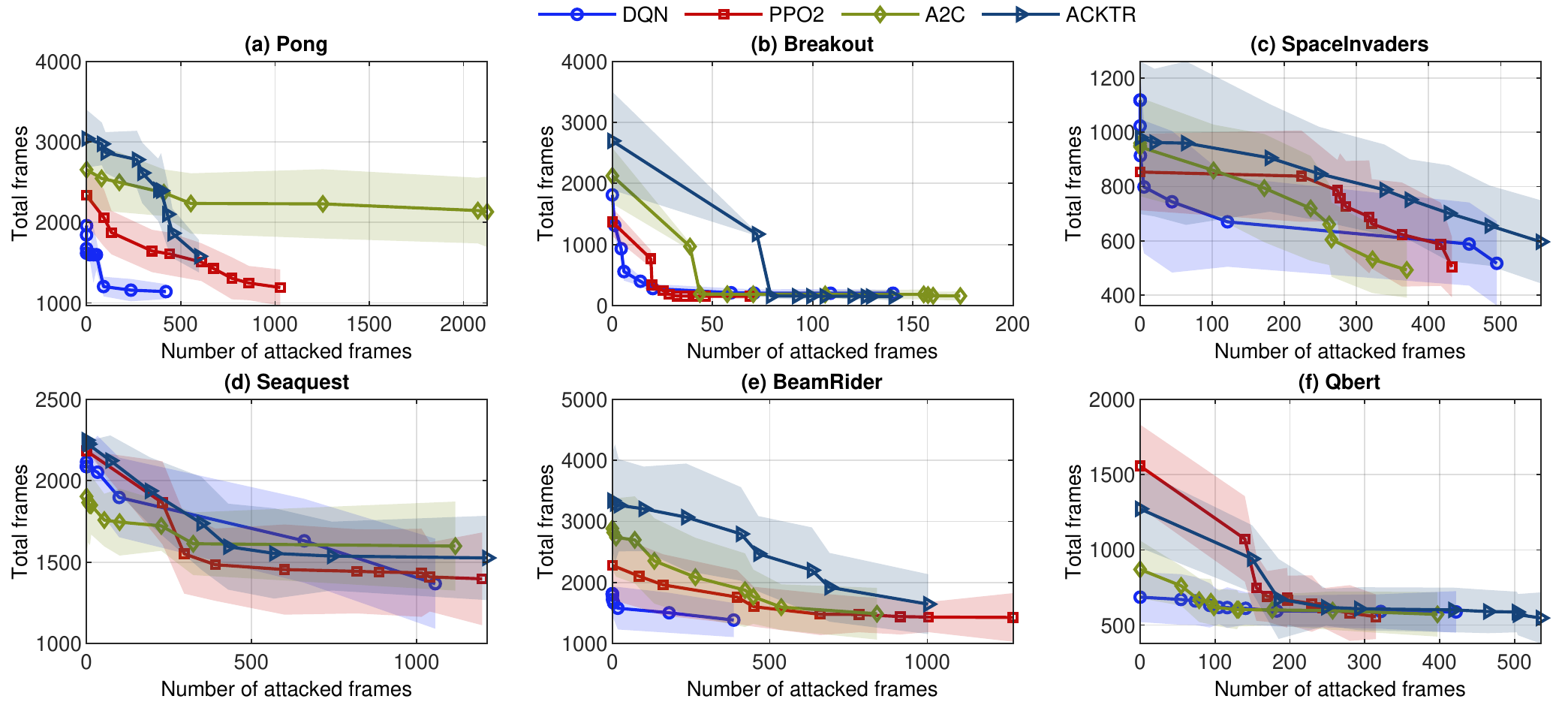}
\caption{The relationship between attacked frames and total frames, where the line and shaded area illustrate the mean and standard deviation of 30 independent runs respectively.}
\label{fig:total_frames_compare_0}
\vspace{-0.1in}
\end{figure*}

To calculate the attack uncertainty for DQN, the soft-max operation is applied to normalize the output.
Moreover, given the image size and pixel value, the  constraints in Eq. (\ref{equ:integer programming}) are set as $\mathcal{I}_\textbf{x}=84$, $\mathcal{I}_\textbf{y}=84$, $\mathcal{I}_\textbf{p}=255$. 
In the GA, the population size $\lambda$, maximum number of objective function evaluations $M^{eval}$, selection rate $\beta$ and mutation rate $\gamma$ are set to 10, 400, 0.2 and 0.1 respectively. However, following Theorem 1, the optimization process is terminated when a positive function value is found.
To investigate the impact of FSA size $n$, we alter its value from 1 to 10 in steps of 1. In other experiments, we analyze the effect of the TCA threshold $\zeta^*$ by setting different values in [0, 1]. This leads to different proportions of attacked frames.
\vspace{-0.1in}
\subsection{Results for Fractional-State Adversary}
We investigate the impact of FSA size $n$, where
$n=1$ corresponds to the extreme case that only one pixel is perturbed. For a fair comparison with respect to different algorithms and games, in the experiments reported in Fig.~\ref{fig:cubesize_comparison} we have set the TCA threshold $\zeta^*$ as the mean of all frames' attack uncertainty values.
Fig.~\ref{fig:cubesize_comparison} reveals the performance (i.e. accumulated reward) drop of the policies trained by the four RL algorithms on the six Atari games. 
Several interesting observations can be noted.

\textbf{(1)} Overall, the FSA size $n$ is positively related to the performance drop. That is, the FSA with larger $n$ is able to deceive the policy to a larger extent.
\textbf{(2)} On most of the games, the policies are almost completely fooled with $n\leq 4$. This indicates a small FSA size is often sufficient to achieve a successful adversarial attack, indicating the efficiency of FSA.
\textbf{(3)} The results indicate that high performance in the absence of attacks is not a guarantee of robustness against adversarial attacks. For instance, in SpaceInvaders, the original performance of A2C is higher than DQN, but the performance of A2C drops faster than that of DQN as shown in Fig.  \ref{fig:cubesize_comparison}(c).
\textbf{(4)} We \textit{surprisingly} find that: on Breakout, with only a single pixel attacked, the policies trained by all the four RL algorithms are deceived, with the accumulated reward dropping by more than 50\%; the visualisation for the single-pixel attack on Breakout is shown in Fig. \ref{fig:breakout comparison}. Such single pixel attack corresponds to an approximate perturbation proportion of $0.01\%$ (i.e., $1/(84 \times 84)$) of the total number of pixels in a frame. On Qbert, the accumulated reward of the policy trained by DQN drops from 10000 to 0 with {two pixels} attacked. 

\subsection{Results for Tactically-Chanced Attack}
We study the tactically-chanced attack (TCA) by altering the TCA threshold $\zeta^*$. Different $\zeta^*$ values lead to different proportions of attacked frames. 
In our exploration, we set the $\zeta^*$ in the range of $(0, 1]$, where $\zeta^*=0$ and $\zeta^*=1$ correspond to $0\%$ and $100\%$ proportion of attacked frames, respectively. Recall that according to the results in Fig.  \ref{fig:cubesize_comparison}, most of the policies are sufficiently fooled with no more than 4 pixels attacked. Hence, for a fair comparison, we set the FSA size $n=4$ in subsequent experiments. 
The results 
are displayed in Fig.  \ref{fig:tca_comparison}, where
we observe that in general the performance (i.e., accumulated reward) decreases for higher TCA threshold $\zeta^*$ values. This demonstrates that the policies are more easily fooled with more frames attacked.  
In addition, we make some other instructive observations.

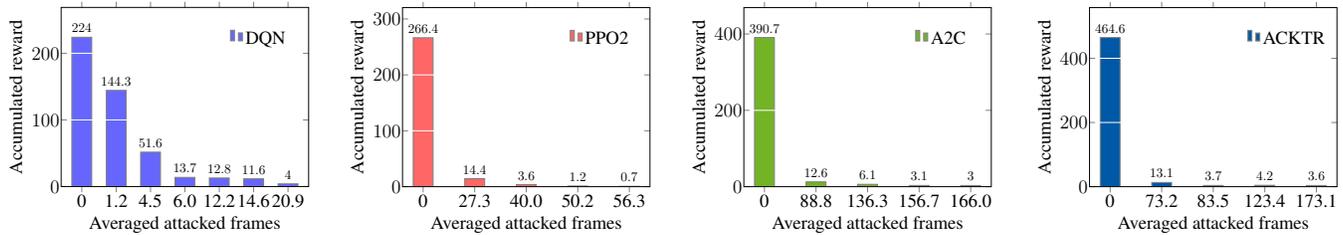
\begin{figure*}[!ht]
\centering
\begin{tikzpicture}[scale=0.5]
\begin{axis}[
ybar, 
width=0.45\textwidth,
height=0.35\textwidth,
enlarge y limits={upper, value=0.2},
enlarge x limits=0.1,
ymin=0,
ymax=224,
ylabel={Accumulated reward},
xlabel={Averaged attacked frames},
ylabel style ={font = \Large},
xlabel style ={font = \Large},
yticklabel style = {font=\Large,xshift=0.5ex},
xticklabel style = {font=\Large,yshift=0.5ex},
bar width=15pt,
legend style={at={(0.8,0.9)}, anchor=north,legend columns= 4, font = \Large},
symbolic x coords={0, 1.2, 4.5, 6.0, 12.2, 14.6, 20.9},
xtick=data,
nodes near coords,
axis on top,
major grid style=white,
ymajorgrids,
legend style={draw=none,/tikz/every even column/.append style={column sep=0.5cm}}]
\addplot[draw=gray,fill=blue!60!white, thick] coordinates {(0,224) (1.2,144.3) (4.5,51.6) (6.0,13.7) (12.2,12.8) (14.6,11.6) (20.9,4.0)};
\legend{DQN}
\end{axis}
\end{tikzpicture}
\hspace{0.1in}
\begin{tikzpicture}[scale=0.5]
\begin{axis}[
ybar, 
width=0.45\textwidth,
height=0.35\textwidth,
enlarge y limits={upper, value=0.2},
enlarge x limits=0.1,
ymin=0,
ymax=267,
ylabel={Accumulated reward},
xlabel={Averaged attacked frames},
ylabel style ={font = \Large},
xlabel style ={font = \Large},
yticklabel style = {font=\Large,xshift=0.5ex},
xticklabel style = {font=\Large,yshift=0.5ex},
bar width=15pt,
legend style={at={(0.8,0.9)}, anchor=north,legend columns= 4, font = \Large},
symbolic x coords={0, 27.3, 40.0, 50.2, 56.3},
xtick=data,
nodes near coords,
axis on top,
major grid style=white,
ymajorgrids,
legend style={draw=none,/tikz/every even column/.append style={column sep=0.5cm}}]
\addplot[draw=gray,fill=red!60!white, thick] coordinates {(0,266.4) (27.3,14.4) (40.0,3.6) (50.2,1.2) (56.3,0.7)};
\legend{PPO2}
\end{axis}
\end{tikzpicture}
\hspace{0.1in}
\begin{tikzpicture}[scale=0.5]
\begin{axis}[
ybar, 
width=0.45\textwidth,
height=0.35\textwidth,
enlarge y limits={upper, value=0.2},
enlarge x limits=0.1,
ymin=0,
ymax=391,
ylabel={Accumulated reward},
xlabel={Averaged attacked frames},
ylabel style ={font = \Large},
xlabel style ={font = \Large},
yticklabel style = {font=\Large,xshift=0.5ex},
xticklabel style = {font=\Large,yshift=0.5ex},
bar width=15pt,
legend style={at={(0.8,0.9)}, anchor=north,legend columns= 4, font = \Large},
symbolic x coords={0, 88.8, 136.3, 156.7, 166.0},
xtick=data,
nodes near coords,
axis on top,
major grid style=white,
ymajorgrids,
legend style={draw=none,/tikz/every even column/.append style={column sep=0.5cm}}]
\addplot[draw=gray,fill=green!40!brown, thick] coordinates {(0,390.7) (88.8,12.6) (136.3,6.1) (156.7,3.1) (166.0,3.0)};
\legend{A2C}
\end{axis}
\end{tikzpicture}
\hspace{0.1in}
\begin{tikzpicture}[scale=0.5]
\begin{axis}[
ybar, 
width=0.45\textwidth,
height=0.35\textwidth,
enlarge y limits={upper, value=0.2},
enlarge x limits=0.1,
ymin=0,
ymax=465,
ylabel={Accumulated reward},
xlabel={Averaged attacked frames},
ylabel style ={font = \Large},
xlabel style ={font = \Large},
yticklabel style = {font=\Large,xshift=0.5ex},
xticklabel style = {font=\Large,yshift=0.5ex},
bar width=15pt,
legend style={at={(0.8,0.9)}, anchor=north,legend columns= 4, font = \Large},
symbolic x coords={0, 73.2, 83.5, 123.4, 173.1},
xtick=data,
nodes near coords,
axis on top,
major grid style=white,
ymajorgrids,
legend style={draw=none,/tikz/every even column/.append style={column sep=0.5cm}}]
\addplot[draw=gray,fill=blue!65!green, thick] coordinates {(0,464.6) (73.2,13.1) (83.5,3.7) (123.4,4.2) (173.1,3.6)};
\legend{ACKTR}
\end{axis}
\end{tikzpicture}
\vspace{-0.05in}
\caption{The impact of attacked frames for accumulated reward on Breakout}
\label{fig:attacktimes_comparison_breakout}
\vspace{-0.2in}
\end{figure*}

\textbf{(1)} In Fig.  \ref{fig:tca_comparison} (a) on Pong and Fig. \ref{fig:tca_comparison} (d) on Seaquest, the policies trained by PPO2 are more sensitive to the increasing of TCA threshold values. In contrast, the performance of policies trained by DQN appears to be more stable with the increasing values of TCA threshold, until it is close to 1. We infer that this observation is because DQN provides Q values instead of directly predicting action probabilities; the resultant increased variance of the derived action distribution leads to much fewer attacked frames under a given threshold value.
\textbf{(2)}
In Fig.  \ref{fig:tca_comparison} (b) on Breakout, the accumulated rewards of the policies trained by PPO2, A2C and ACKTR decrease to 0 when the TCA threshold value is higher than 0.3. This indicate that a small TCA threshold value is enough to completely deceive those policies. 
\textbf{(3)} In Fig.  \ref{fig:tca_comparison} (f) on Qbert, when the TCA threshold value is higher than 0.6, all policies can only obtain a reward less than 2000. Especially for the policy trained by PPO2, the accumulated reward drops from 14000 to 2000.

We further analyze the exact number of attacked frames with different TCA threshold values. As shown in Fig. \ref{fig:attacked_frames_compare_0}, the number of attacked frames naturally increases with the prescribed threshold value. In Fig. \ref{fig:attacked_frames_compare_0} (b), on Breakout, when the TCA threshold value is smaller than 0.2, less than 50 frames are attacked. A similar finding is also observed on Beamrider, where less than 100 frames are attacked when the TCA threshold value is smaller than 0.2. 

We also analyze the relationship between the number of attacked frames and the number of total frames in Fig. \ref{fig:total_frames_compare_0}.  We find similar trends on the six games, where the the total number of frames in the game decreases when more frames are attacked. This observation is particularly apparent for Breakout in Fig. \ref{fig:total_frames_compare_0} (b), where attacking less than 50 frames considerably reduces the total number of frames. This results from the fact that the attacked frames may cause (i) termination of the game, or (ii) a loss of life. Both outcomes greatly shorten the episode, resulting in a smaller number of total frames. For instance, if the paddle in Breakout is deceived to miss the ball, the agent would lose 1/5 life.

To explore a more restrictive setting, we additionally examine the single-pixel ($n=1$) attack on limited frames. We illustrate the results on Breakout in Fig.  \ref{fig:attacktimes_comparison_breakout}, and similar trends can be observed on other games as well.  
From this figure, we find that by only attacking six frames on average, the policy trained by DQN is totally fooled with reward decreasing from 224 to 13.7. Here, six frames correspond to an attack ratio of $6/535\approx 1\%$, where 535 is the averaged number of total frames. Similar findings are also observed on policies trained by PPO2, A2C and ACKTR. For instance, the policy trained by PPO2 shows a significant reward decline with only 27 frames attacked under the same TCA setting as other policies. 

\begin{figure*}
\captionsetup[subfigure]{labelformat=empty}
\centering
\begin{subfigure}{0.245\textwidth}
    \centering
     \includegraphics[width=\textwidth]{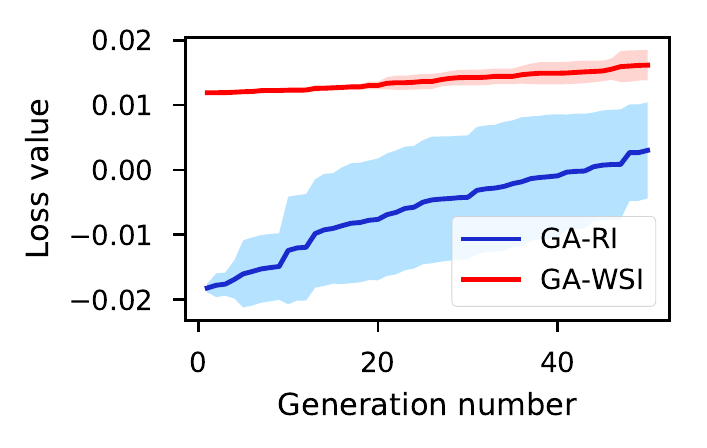}
     \vspace{-0.2in}
     \caption{Pong: $\xi_1$}
\end{subfigure}
\begin{subfigure}{0.245\textwidth}
    \centering
     \includegraphics[width=\textwidth]{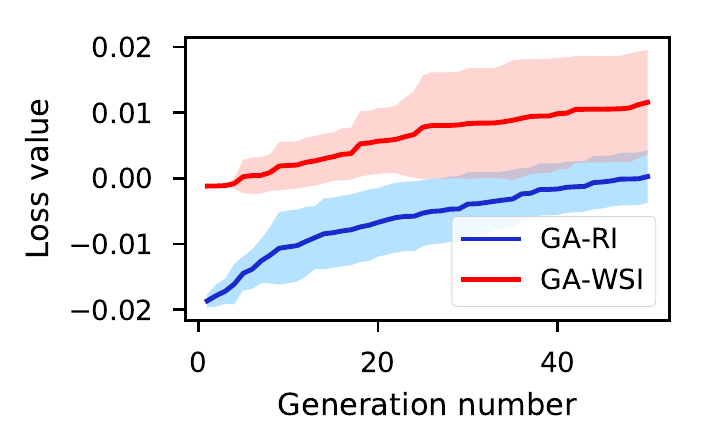}
     \vspace{-0.2in}
     \caption{Pong: $\xi_2$}
\end{subfigure}
\begin{subfigure}{0.245\textwidth}
    \centering
     \includegraphics[width=\textwidth]{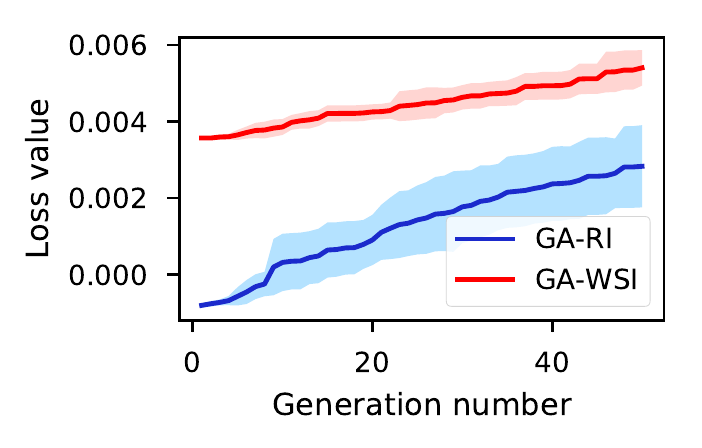}
     \vspace{-0.2in}
     \caption{Pong: $\xi_3$}
\end{subfigure}
\begin{subfigure}{0.245\textwidth}
    \centering
     \includegraphics[width=\textwidth]{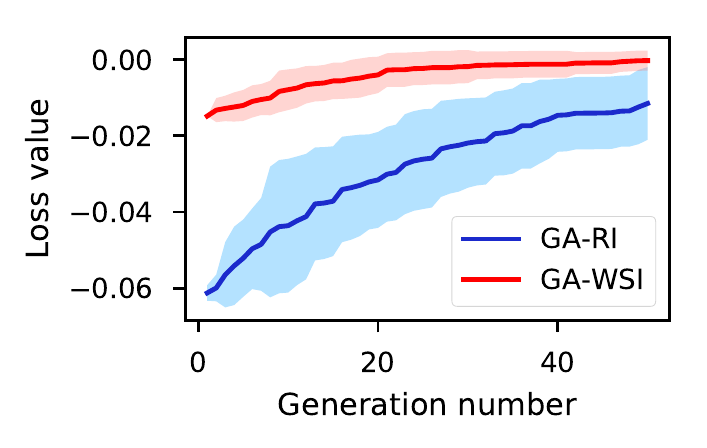}
     \vspace{-0.2in}
     \caption{Pong: $\xi_4$}
\end{subfigure}\\
\begin{subfigure}{0.245\textwidth}
    \centering
     \includegraphics[width=\textwidth]{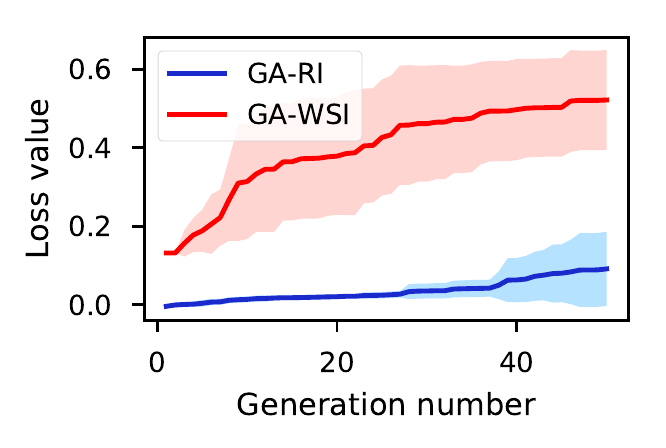}
     \vspace{-0.2in}
     \caption{Breakout: $\xi_1$}
\end{subfigure}
\begin{subfigure}{0.245\textwidth}
    \centering
     \includegraphics[width=\textwidth]{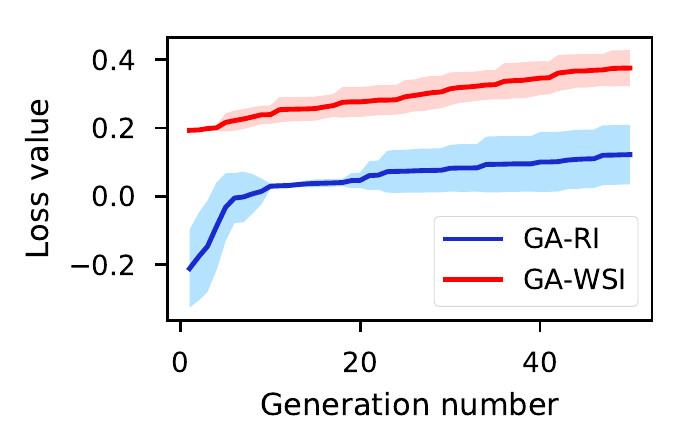}
     \vspace{-0.2in}
     \caption{Breakout: $\xi_2$}
\end{subfigure}
\begin{subfigure}{0.245\textwidth}
    \centering
     \includegraphics[width=\textwidth]{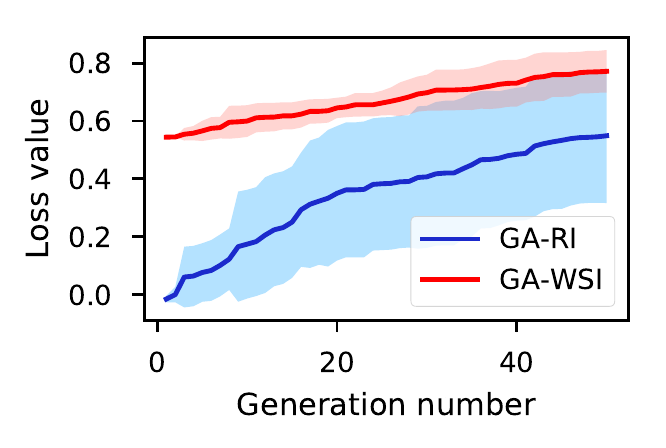}
     \vspace{-0.2in}
     \caption{Breakout: $\xi_3$}
\end{subfigure}
\begin{subfigure}{0.245\textwidth}
    \centering
     \includegraphics[width=\textwidth]{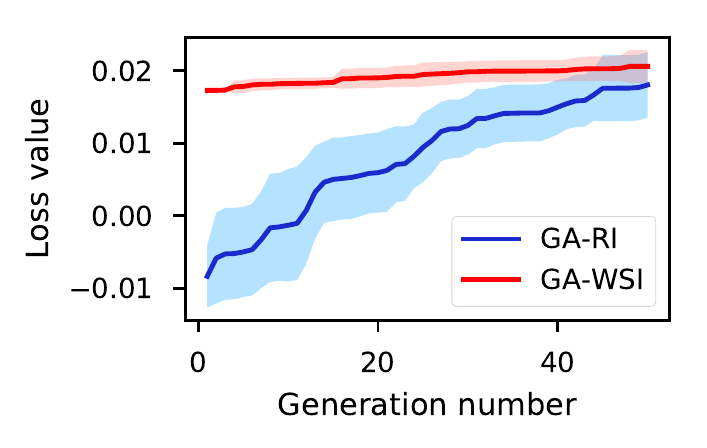}
     \vspace{-0.2in}
     \caption{Breakout: $\xi_4$}
\end{subfigure}\\
\begin{subfigure}{0.245\textwidth}
    \centering
     \includegraphics[width=\textwidth]{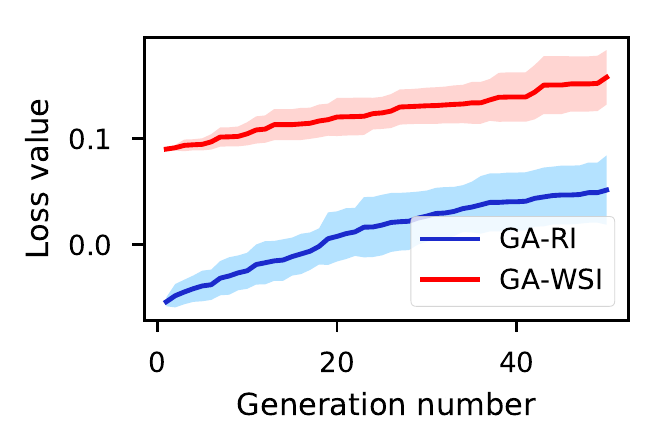}
     \vspace{-0.2in}
     \caption{Space Invaders: $\xi_1$}
\end{subfigure}
\begin{subfigure}{0.245\textwidth}
    \centering
     \includegraphics[width=\textwidth]{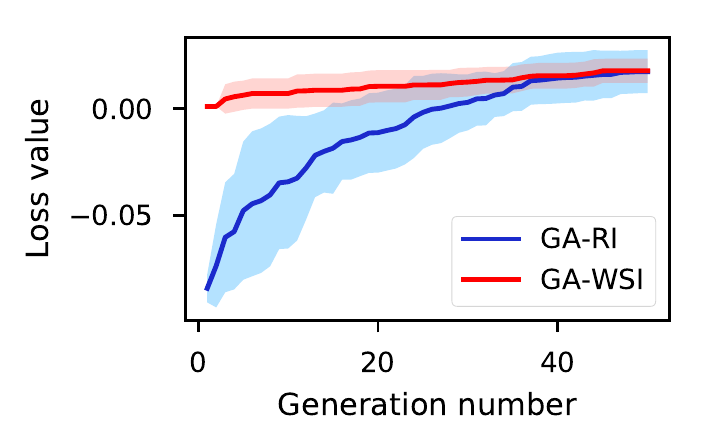}
     \vspace{-0.2in}
     \caption{Space Invaders: $\xi_2$}
\end{subfigure}
\begin{subfigure}{0.245\textwidth}
    \centering
     \includegraphics[width=\textwidth]{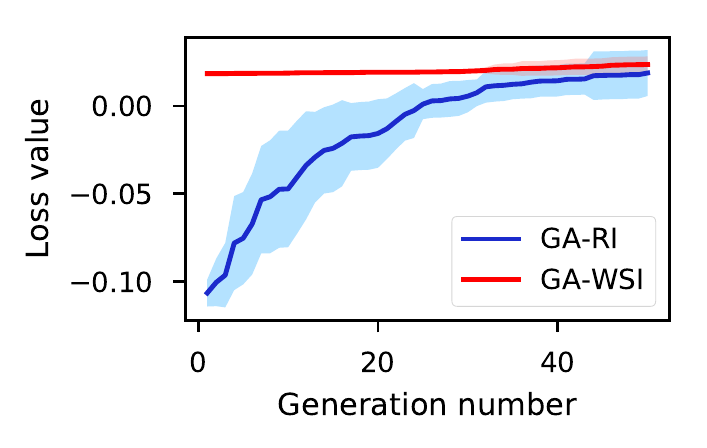}
     \vspace{-0.2in}
     \caption{Space Invaders: $\xi_3$}
\end{subfigure}
\begin{subfigure}{0.245\textwidth}
    \centering
     \includegraphics[width=\textwidth]{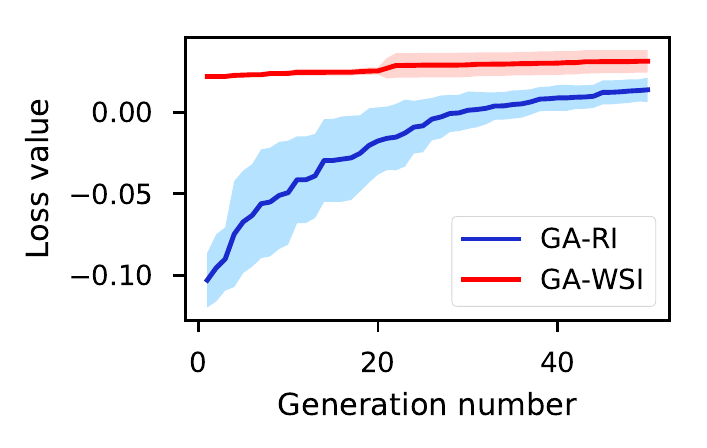}
     \vspace{-0.2in}
     \caption{Space Invaders: $\xi_4$}
\end{subfigure}\\
\begin{subfigure}{0.245\textwidth}
    \centering
     \includegraphics[width=\textwidth]{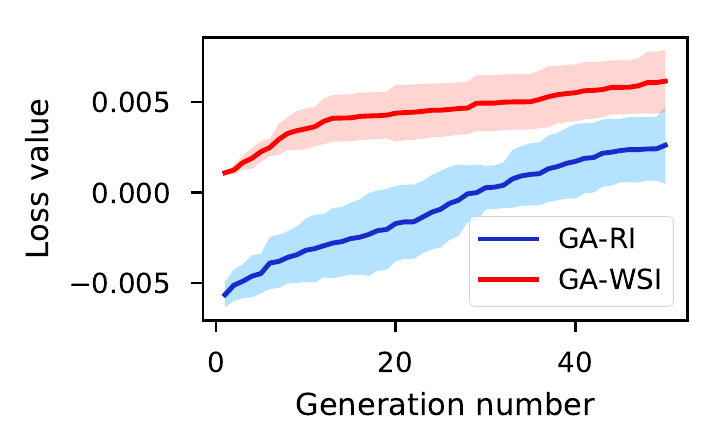}
     \vspace{-0.2in}
     \caption{Seaquest: $\xi_1$}
\end{subfigure}
\begin{subfigure}{0.245\textwidth}
    \centering
     \includegraphics[width=\textwidth]{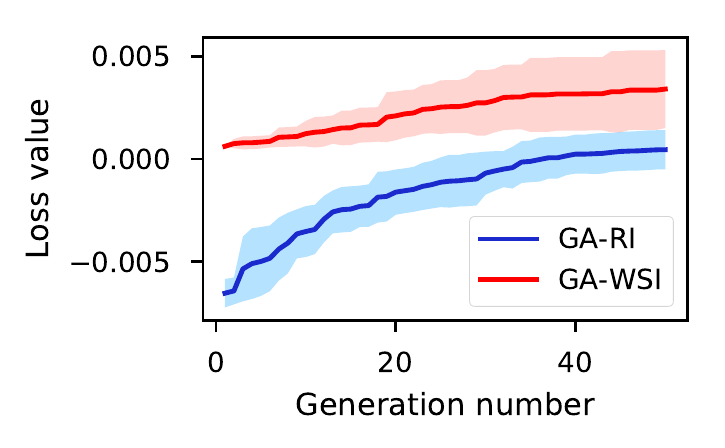}
     \vspace{-0.2in}
     \caption{Seaquest: $\xi_2$}
\end{subfigure}
\begin{subfigure}{0.245\textwidth}
    \centering
     \includegraphics[width=\textwidth]{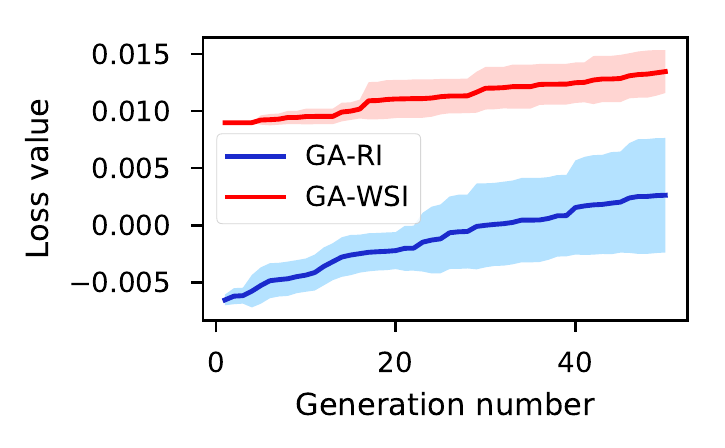}
     \vspace{-0.2in}
     \caption{Seaquest: $\xi_3$}
\end{subfigure}
\begin{subfigure}{0.245\textwidth}
    \centering
     \includegraphics[width=\textwidth]{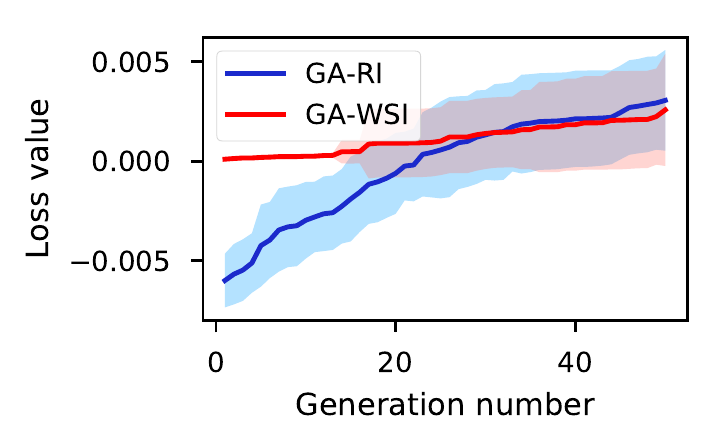}
     \vspace{-0.2in}
     \caption{Seaquest: $\xi_4$}
\end{subfigure}\\
\begin{subfigure}{0.245\textwidth}
    \centering
     \includegraphics[width=\textwidth]{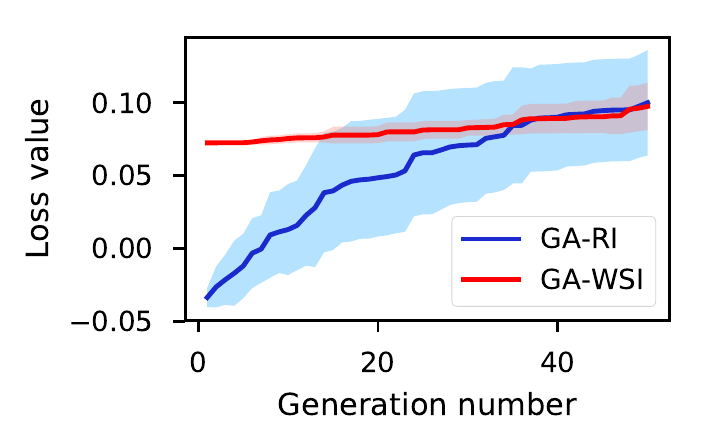}
     \vspace{-0.2in}
     \caption{Qbert: $\xi_1$}
\end{subfigure}
\begin{subfigure}{0.245\textwidth}
    \centering
     \includegraphics[width=\textwidth]{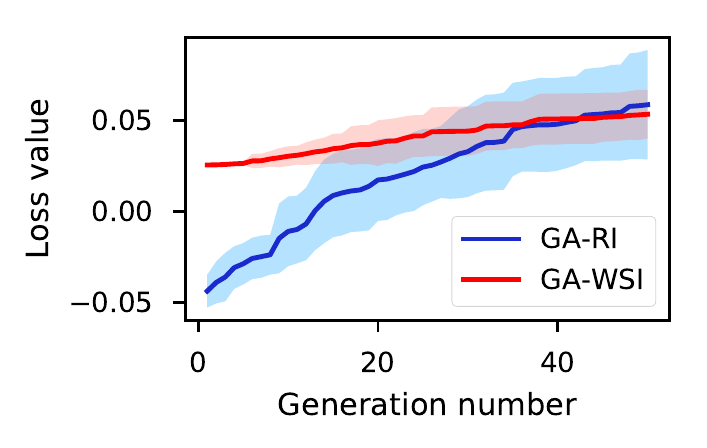}
     \vspace{-0.2in}
     \caption{Qbert: $\xi_2$}
\end{subfigure}
\begin{subfigure}{0.245\textwidth}
    \centering
     \includegraphics[width=\textwidth]{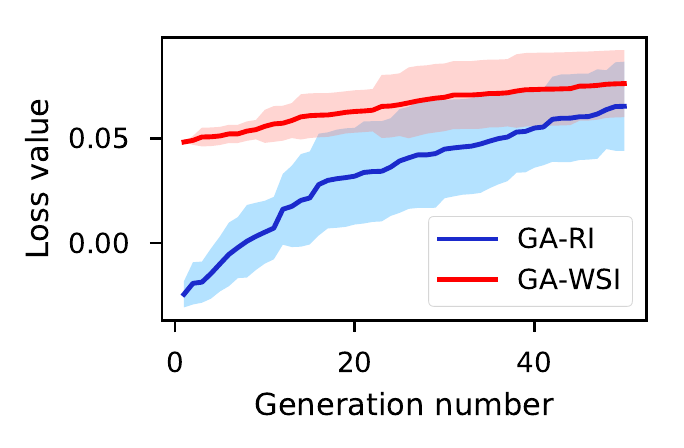}
     \vspace{-0.2in}
     \caption{Qbert: $\xi_3$}
\end{subfigure}
\begin{subfigure}{0.245\textwidth}
    \centering
     \includegraphics[width=\textwidth]{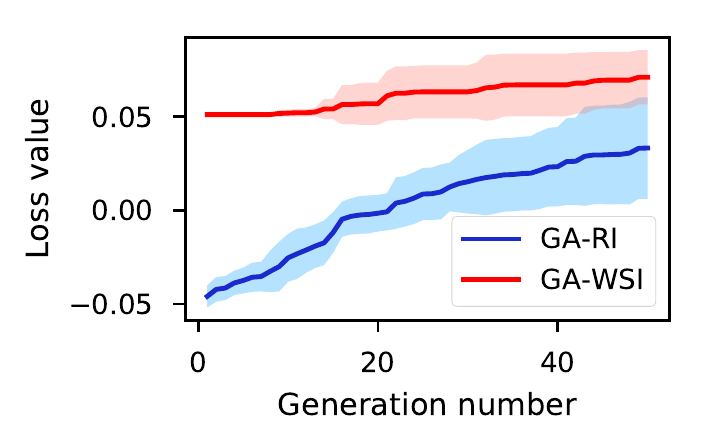}
     \vspace{-0.2in}
     \caption{Qbert: $\xi_4$}
\end{subfigure}\\
\begin{subfigure}{0.245\textwidth}
    \centering
     \includegraphics[width=\textwidth]{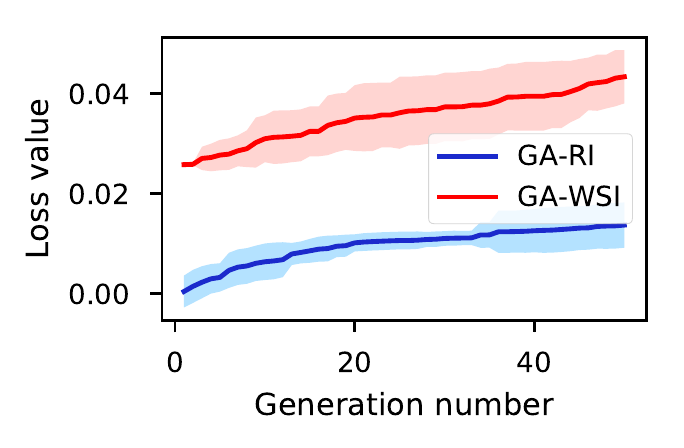}
     \vspace{-0.2in}
     \caption{Beamrider: $\xi_1$}
\end{subfigure}
\begin{subfigure}{0.245\textwidth}
    \centering
     \includegraphics[width=\textwidth]{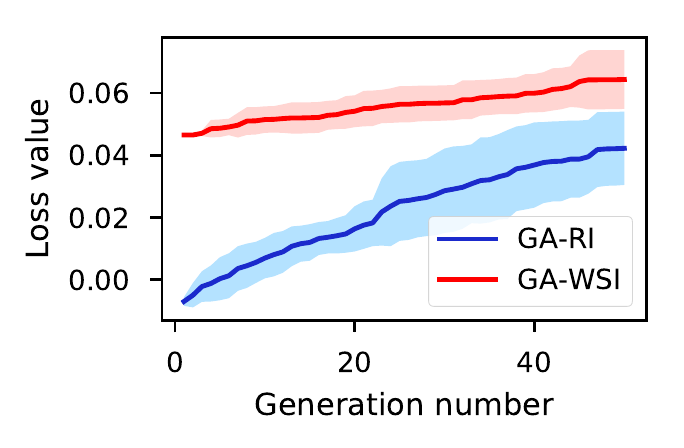}
     \vspace{-0.2in}
     \caption{Beamrider: $\xi_2$}
\end{subfigure}
\begin{subfigure}{0.245\textwidth}
    \centering
     \includegraphics[width=\textwidth]{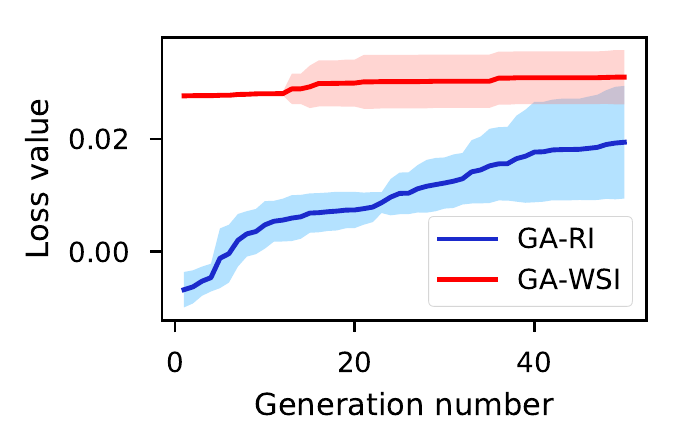}
     \vspace{-0.2in}
     \caption{Beamrider: $\xi_3$}
\end{subfigure}
\begin{subfigure}{0.245\textwidth}
    \centering
     \includegraphics[width=\textwidth]{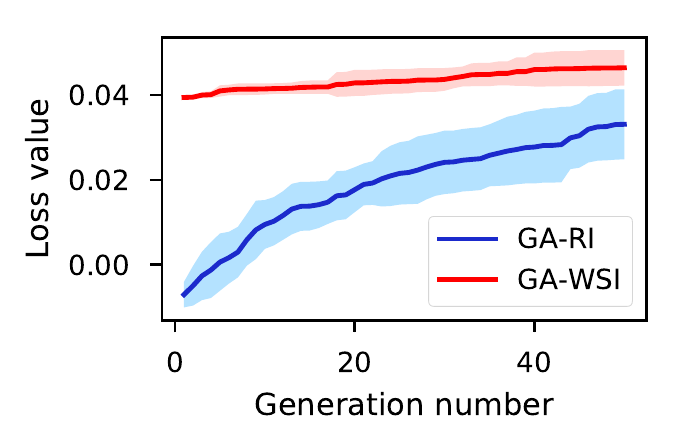}
     \vspace{-0.2in}
     \caption{Beamrider: $\xi_4$}
\end{subfigure}\\
\caption{The comparison between random initialization (GA-RI) and warm start initialization (GA-WSI), where the line and shaded area illustrate the mean and standard deviation of 10 independent runs respectively.}
\label{fig:Rand_Warm}
\end{figure*}
\subsection{Attack Initialization Analysis}
The GA population initialization plays an important role in efficiently generating successful adversarial examples, and thus highlights the practicality of the approach. To this end, we study the effects of random initialization versus warm starting based on prior data. In RL, the simplest form of prior data originates from the correlation between states in a Markov Decision Process (MDP); this also distinguishes RL from other supervised learning tasks (e.g., classification), where the predictions on different data points are independent. 
Thus, with the existence of correlations between adjacent frames in mind, we set up different types of population initializations in the GA: (1) \textbf{GA-RI}: random initialization (RI) of $pop_0$; (2) \textbf{GA-WSI}: warm start initialization (WSI) of $pop_0$ by inserting the optimal adversary $x_p^*$ found in the previous attacked frame.

We randomly attack four frames for all the six games. On each attacked frame, the GA optimization is executed 10 times independently on both types of initialization.
The comparison results are shown in Fig. \ref{fig:Rand_Warm}.
In general, we clearly see that GA-WSI significantly speeds up the optimization convergence in finding successful adversarial examples (i.e., corresponding to positive objective function values). In particular, on many frames (e.g., Pong: $\xi_1$, $\xi_2$, $\xi_4$; Breakout: $\xi_2$; Space Invaders: $\xi_3$; Qbert: $\xi_3$, $\xi_4$), the GA-RI takes significantly more time (or is even unable) to obtain a positive objective value; whereas GA-WSI achieves it easily.

\emph{We interestingly find that, in some cases, GA-WSI provides a successful adversary without the need of any optimization.
This observation underlines the real possibility of launching successful adversarial attacks on deep RL policies in \textbf{real-time} at little/no computational cost, a threat that raises severe concerns about their deployment especially in safety critical applications}. As such, it points to two critical directions for future research. On the one hand, there is a need for more comprehensive exploration of procedures of learning from experiential frames, that could lead to the generation of potent adversaries on-the-fly. On the other hand, a significant void in the trustworthiness of existing RL policies is revealed, one that could possibly be addressed through more robust training algorithms.

\section*{Conclusion}
This paper explores \textit{minimalistic scenarios} for adversarial attack in deep reinforcement learning (RL), comprising (1) black-box policy access where the attacker only has access to the input (state) and output (action probability) of an RL policy;
(2) fractional-state adversary where only a small number of pixels are perturbed, with the extreme case of one-pixel attack; and (3) tactically-chanced attack where only some significant frames are selected to be attacked. We verify these settings on policies that are trained by four state-of-the-art RL algorithms on six Atari games. We surprisingly find that: (i) with only \textbf{\textit{a single pixel}} ($\approx0.01\%$ of the state) attacked, the trained policies can be significantly deceived; and (ii) by merely attacking around {$1\%$ frames}, the policy trained by DQN is totally fooled on certain games. Our immediate future work will focus on further investigating different initialization schemes that could better exploit the correlations across different attacked frames. Our goal is to thoroughly analyze the plausibility of generating adversarial attacks in real-time for time-sensitive RL applications.

\section*{Acknowledgement}
This work is funded by the National Research Foundation, Singapore under its AI Singapore programme [Award No.: AISG-RP-2018-004] and the Data Science and Artificial Intelligence Research Center (DSAIR) at Nanyang Technological University. Any opinions, findings and conclusions or recommendations expressed in this material are those of the authors and do not reflect the views of National Research Foundation, Singapore.

\bibliography{reference}
\bibliographystyle{IEEEtran}
\vspace{-0.2in}
\newpage
\begin{IEEEbiography}[{\includegraphics[width=1in,height=1.25in,clip,keepaspectratio]{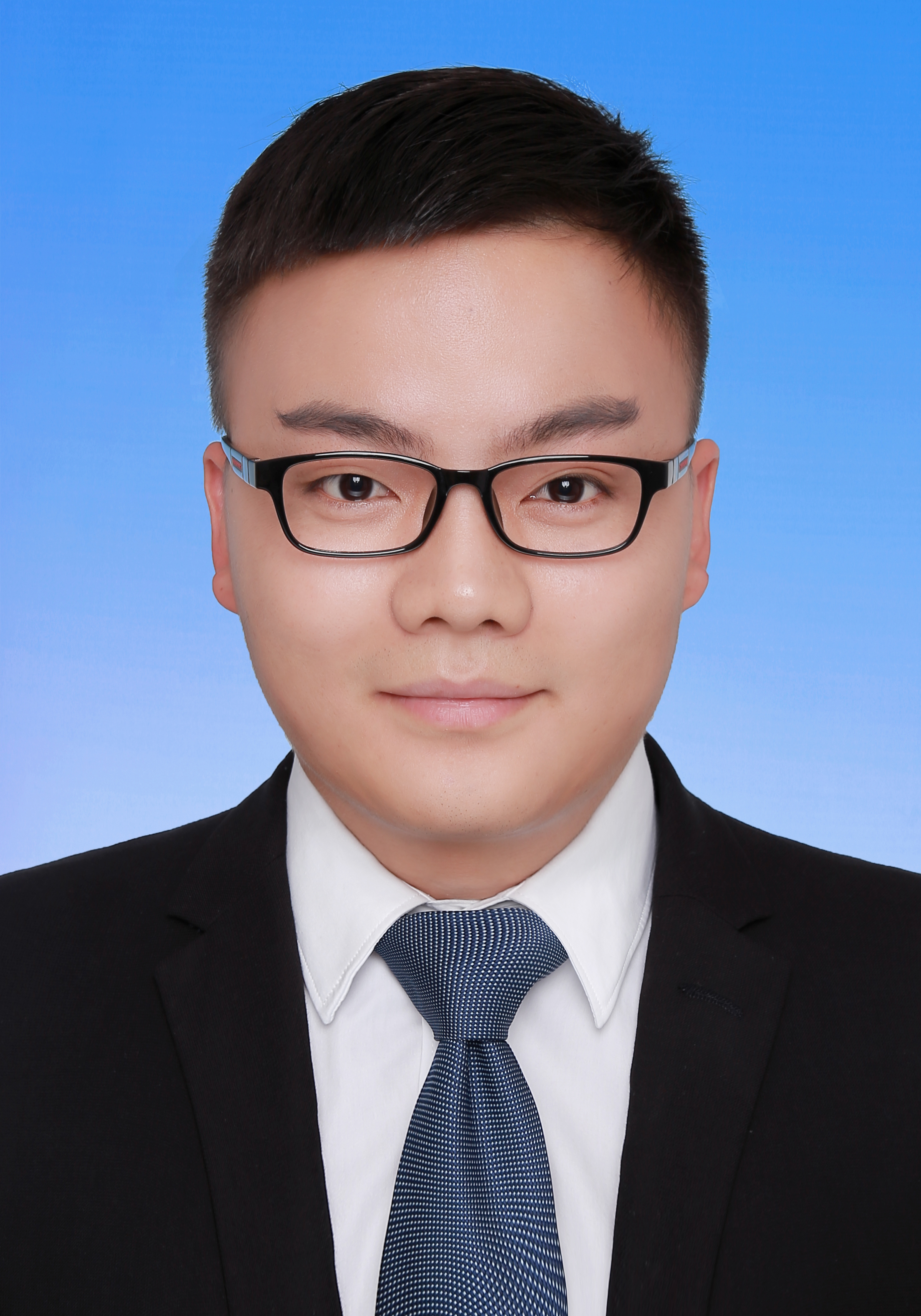}}]{Xinghua Qu}
received the B.S. degree in Aircraft Design and Engineering
from Northwestern Polytechnic University, China, in
2014, and the M.S. degree in Aerospace Engineering, School of Astronautics, Beihang
University, China, in 2017. He is currently the Phd student from school of computer science and engineering, Nanyang Technological University, Singapore. His current research focus includes reinforcement learning and adversarial machine learning.
\end{IEEEbiography}
\vspace{-0.4in}
\begin{IEEEbiography}[{\includegraphics[width=1in,height=1.25in,clip,keepaspectratio]{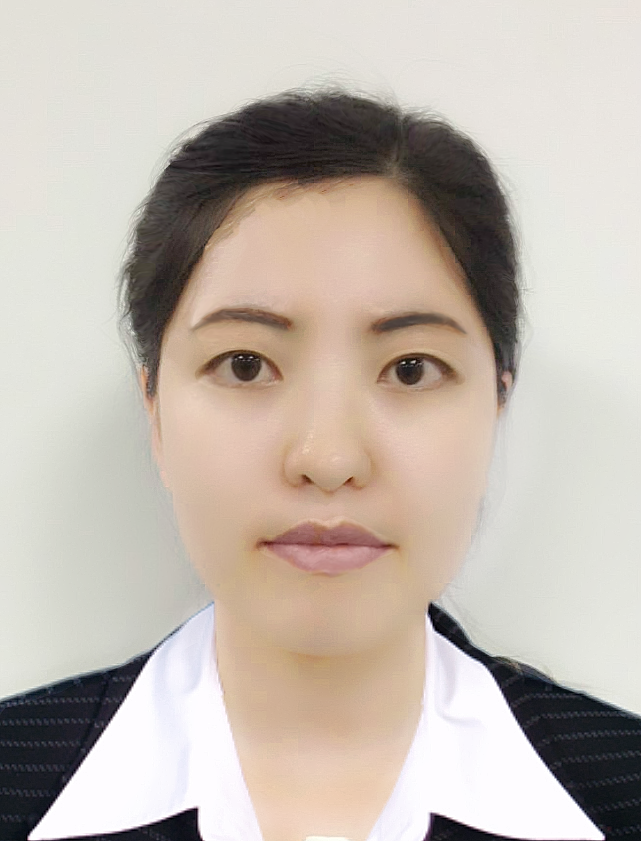}}]{Zhu Sun}
received her Ph.D. degree from School of Computer Science and Engineering,
Nanyang Technological University (NTU), Singapore, in 2018. During her Ph.D.
study, she focused on designing efficient recommendation algorithms by considering side information. Her research has been published in leading
conferences and journals in related domains (e.g., IJCAI, AAAI, CIKM, 
ACM RecSys and ACM UMAP,).
Currently, she is a research
fellow at School of Electrical and Electronic Engineering, NTU, Singapore.
\end{IEEEbiography}
\vspace{-0.4in}
\begin{IEEEbiography}[{\includegraphics[width=1in,height=1in,clip,keepaspectratio]{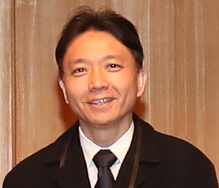}}]{Yew-Soon Ong (M’99-SM’12-F’18)}
received a Ph.D. degree on artificial intelligence in complex design from the University of Southampton, U.K., in 2003. He is a President’s Chair Professor in
Computer Science at the Nanyang Technological University (NTU), and holds the position of Chief Artificial Intelligence Scientist at the Agency for Science, Technology and Research Singapore. At
NTU, he currently serves as Director of the Data Science and Artificial Intelligence Research Center and Director of the Singtel-NTU Cognitive \& Artificial Intelligence Joint Lab.
His current research focus is in artificial and computational intelligence. He is founding Editor-in-Chief of the IEEE Transactions on Emerging Topics in Computational Intelligence and Associate
Editor of several IEEE Transactions. He has received several IEEE outstanding paper awards, listed as a Thomson Reuters highly cited researcher and among the World’s Most Influential Scientific Minds.
\end{IEEEbiography}
\vspace{-0.4in}
\begin{IEEEbiography}[{\includegraphics[width=1in,height=1.25in,clip,keepaspectratio]{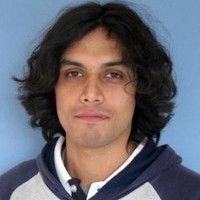}}]{Abhishek Gupta}
received his PhD in Engineering Science from the University of Auckland, New
Zealand, in the year 2014. He currently serves as a Scientist in the
Singapore Institute of Manufacturing Technology
(SIMTech), at the Agency for Science, Technology and
Research (A*STAR), Singapore. He has diverse
research experience in the field of computational
science, ranging from numerical methods in engineering physics, to topics in computational intelligence. His current research focus is in the development of memetic computation and probabilistic model-based algorithms for automated knowledge extraction and transfer across optimization problems, with diverse applications spanning cyber physical production systems and engineering design.
\end{IEEEbiography}
\vspace{-0.4in}
\begin{IEEEbiography}[{\includegraphics[width=1in,height=1.25in,clip,keepaspectratio]{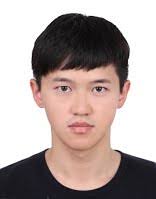}}]{Pengfei Wei}
received his Ph.D. Degree from School of Computer Science and Engineering,
Nanyang Technological University (NTU), Singapore, in 2018. During his Ph.D.
study, he focused on domain adaptation and gaussian processes. His research has been published in leading
conferences and journals in related domains (e.g., ICML, IJCAI, ICDM, IEEE TNNLS, IEEE TKDE).  Currently, he is a research
fellow at School of Computing, National University of Singapore.
\end{IEEEbiography}

\end{document}